MSc Computational Science
Master Thesis

# Learning a Representation Map for Robot Navigation using Deep Variational Autoencoder

by

Kaixin Hu

April 2018

*Supervisor:*
Peter O'Connor

*Assessor:*
dhr. dr. E. (Stratis) Gavves
Dr. Zeynep Akata

University of Amsterdam

# Acknowledgments

I would like to thank my daily-supervisor Peter O'Connor for promoting this interesting project. Also I very appreciate his continuous support and encouragement: from providing necessary resources for this work to checking the schedule and giving suggestions every week. Thanks for his guidance on this project and freedom given to try methods on my own, I learned a lot from this project. It is many discussions on the methods and problems that lead to this final work. I would also like to thank Changyong Oh, Shuai Liao, Matthias Reisser, etc. in the Machine Learning Lab who helped me with technical problems when Peter was away on business.

I would like to thank dhr. dr. E. (Stratis) Gavves for giving the valuable suggestions on improvement of the results and being part of my thesis defense committee. I would also like to thank Dr. Zeynep Akata for reading this work and being part of my thesis defense committee.

Finally, I would like to thank my family and friends at home and in Amsterdam for their support and accompany. Especially, I would like to dedicate this thesis to my parents for their utmost support and faith in me.

# Abstract

The aim of this work is to use Variational Autoencoder (VAE) to learn a representation of an indoor environment that can be used for robot navigation.

We use images extracted from a video, in which a camera takes a tour around a house, for training the VAE model with a 4 dimensional latent space. After the model is trained on the images, each real frame has a corresponding representation point on manifold in the latent space, and each representation point has corresponding reconstructed image. For the navigation problem, we map the starting image and destination image to the latent space, then optimize a path on the learned manifold connecting the two points, and finally map the path back through decoder to a sequence of images. The ideal sequence of images should correspond to a route that is spatially continuous - i.e. neighbor images in the route should correspond to neighbor locations in physical space. Such a route could be used for navigation with computer vision techniques, i.e. a robot could follow the image sequence from starting location to destination in the environment step by step. We implement this algorithm, but find in our experimental results that the resulting route is not satisfactory. The routes produced in the experiments are not ideal, for they consist of several discontinuous image frames along the ideal routes, so that the route could not be followed by a robot with computer vision techniques in practice. In our evaluation, we propose two reasons for our failure to automatically find continuous routes: (1) The VAE tends to capture global structures, but discard the details; (2) the Euclidean similarity metric used for measuring continuity between house images is sub-optimal. For further work, we propose: trying other generative models like VAE-GANs which may be better at reconstructing the details to learn the representation map, and adjusting the similarity metric in the path selecting algorithm.

# Contents

# Chapter 1

# Introduction

## 1.1 Problem statement

The aim of this work is to use Variational Autoencoder (VAE) to learn a representation of an indoor environment that can be used for robot navigation.

Specifically, we toured through a house, recording a video of the tour, which should cover the environment as completely as possible. The video is used as training set for both VAE map building in Chapter 4 and later with computer vision techniques to navigate the robot under the a produced path in the latent space map in Chapter 6. The training data consists of 10000 frames of image extracted from the video. After training the VAE model with these data, we get a 4-D learned latent manifold, w.r.t the training data. This means that for any camera image in that environment, if fed through the learned encoder, we get a corresponding latent variable, i.e.4-D vector; and then if the latent variable is fed through the decoder, we get a reconstruction of the original image. We consider that the learned map captures all the useful spatial information if it has two properties[1]: continuity, i.e. the neighbour points on the manifold represent real frames that also are neighbourhood geographically; completeness, i.e. similar real frames will have different representations in the latent space, so each real frame can be recognized and distinguished by the model.

With the hypothesis that the learned map is continuous and complete, we design experiments to produce route given starting and ending locations. Given two camera images, i.e. a starting one and a destination one, there are corresponding points $z_s$ and $z_d$ in the latent space. The robot can select a path[2] on the latent map connecting

[1]This definitions have nothing to do with mathematical properties of manifold in geometry

[2]For the definition of path, please refer to Chapter 1.3

$z_s$ and $z_d$. We choose a geodesic[3], which is the shortest path in a manifold, as the path. If the learned map is continuous and complete, there should be a reasonable and continuous route[4] with respect to the geodesic. We consider the navigation to be successful if the route produced is continuous and reasonable, i.e. the route consists of a sequence of images that a robot could plausibly follow to get from the location of the starting image to the location of the destination image. By continuity, we mean that neighbour images in the route correspond to nearby locations, and the transformations between these images correspond to a simple perspective transform or shift, which can be figured out with computer techniques.

Our experiments (Chapter 6) show that the produced routes are not ideal for they consist of several discontinuous image frames along the ideal routes, which means not able to be figured out by computer vision techniques. This indicates that the hypothesis of continuity and completeness of the learned map is not true. Our evaluations of the experiments indicate that the reason may due to the VAE model's inability to capture the details in training images and the Euclidean similarity metric which is sub-optimal for measuring distance in house images.

## 1.2 Background information

### 1.2.1 Robot navigation

Robot navigation is a core problem in Robotics. [DeSouza and Kak, 2002] summarized the various sub-fields of robot navigation. There are two major research fields: Indoor and outdoor navigation. The field of indoor navigation has three approached: Mapless Navigation, Map-Based Navigation and Map-Building-Based Navigation. In mapless Navigation, systems recognize objects in the environment or track those objects by generating motions, by optical flows[Santos-Victor et al., 1993], appearance-based paradigm[Gaussier et al., 1997][Joulain et al., 1997], or recognition of specific objects [Kim and Nevatia, 1999][Kim and Nevatia, 1998]. To implement map based or map-building navigation, there are three interdependent process[Levitt and Lawton, 1990][Meyer and Filliat, 2003]: Map-learning, Localization, and Path-planning. Specifically, the processes are building a map of the unknown environment and simultaneously determining the position of itself in the environment, moving itself through[Fuentes-Pacheco et al., 2015]. Map-based or map-building based-systems construct geometric/metric or topological framework of the environment. In a topological framework, the environment is represented by a set of distinctive places[Kuipers and Byun, 1991], and by the way a robot can go from one place to another[Filliat and Meyer, 2003]. In metric framework, the environment is represented as a set of objects with coordinates in a 2D space, and they use idiothetic[5] information to directly monitor the robot's position in the space. For detail various method on this topic can refer to [Filliat and Meyer, 2003] and [Meyer and Filliat, 2003]. There are also some work[Arleo and Gerstner, 2000] in the cognitive field using multi-layer neural architecture to learn a vision representation input of the environment, and together with idiothetic representation and allothetic representation yeilds a stable state space representation. But their aim is to use robot to study the role of hippocampus in spatial cognition and navigation.

[3]For the definition of geodesic, please refer to Chapter 1.3

[4]For the definition of route, please refer to Chapter 1.3

In this work we try to solve the SLAM problem by learning a representation map using a latent variable model,i.e. a variational autoencoder. If successful, we hope that given two locations we can use the learned map to produce a path quickly with no need for other training data any more.

### 1.2.2 Representation learning

Representation learning is essential in machine learning, [Bengio et al., 2013] pointed out it is because representation learning can be convenient to express many general priors about the world around us. One reason is the manifold hypothesis: while in its raw representation such data may appear to live in a high dimensional space, in reality its probability density is likely to be relatively high only along stripes of a much lower-dimensional non-linear sub-manifold, embedded in this high-dimensional Euclidean space[Rifai et al., 2011]. [Lin and Zha, 2008] provided a summary of manifold learning method. Linear manifold learning method, i.e. PCA(principle component analysis), MDS(multidimensional scaling), LDA(linear discriminant analysis), can only deal with the flat Euclidean structures. Non-linear extensions of PCA and MDS, i.e. SOMs, principal curve, autoencoder, and GTMs(generative topographic maps), suffer from cost function designing and too many free parameters. [Lin and Zha, 2008] also mentioned kernel based extensions of this methods, i.e. KPCA and KFD, and a large number of nonlinear manifold learning, i.e. ISOMAP(isometric feature learning), LLE(locally linear embedding), Laplacian eigenmaps, Hessian eigenmaps, and so on.

Variational Autoencoders proposed by [Kingma and Welling, 2013] and [Rezende

[5]By [Filliat and Meyer, 2003], Idiothetic source provides internal information about the robot's movements. Allothetic source provides external information about the environment.

et al., 2014] in parallel are generative models that learn the data structure. They use a learned approximate inference model to estimate the posterior over the latent variables given the data. The implementation algorithm of VAE takes a form of auto-encoder, and make use of the back-propagation and gradient descent method. Here is a brief introduction to auto-encoder, more detail one can be found in following chapters. Given a set of input (x), i.e. a set of images, then the model learns two functions: encoder and decoder. z=encode(x) takes an input x, and transfers it into a point in the "latent space" Z. x = decode(z) takes a point in the latent space z and transforms it into an image. So it is also called reconstructed model. The learning process is to lower the reconstruction error. Because this can lead to overfitting, different forms of regularized auto-encoder are used: contractive auto-encoder[Rifai et al., 2011], sparse auto-encoder[Poultney et al., 2007], denoising auto-encoder[Vincent et al., 2010]. VAE is a form of autoencoder that maximizes lower bound on the probability of the data. Thus VAE has a regularization term that is the KL divergence between the prior and the posterior. VAE makes a connection between auto-encoders, directed probabilistic models and stochastic variational inference.

## 1.3 Definition of terms

### 1.3.1 Map/Representation/Latent space structure/Manifold

In this work, we use the terms "map", "representation" "latent space structure", and "manifold" interchangeably.

For more detailed and step-by-step mathematical definitions on manifold, path, curve, and geodesics in the context of metric geometry, please refer to [Persson, 2013]. Here, we introduce it briefly. A manifold locally resembles a subset of the n-dimensional Euclidean space and a Riemannian metric defines concepts like angles, volumes, length of curves, etc. Together these object constitutes a Riemannian manifold in Riemannian geometry. [Persson, 2013] gave a visual explanation of what Riemannian geometry: the length of a path a hiker traverses in a mountain region. In Euclidean geometry, planning the path is independent of the terrain, You simply connect the two positions and project the straight line onto the terrain, thus maybe resulting in a path very tiresome. While in Riemannian geometry, the Riemannian metric would govern the length of the paths by "penalizing" paths with rough terrain by adjusting the length path by not only the length of the path in the Euclidean sense but also how tiresome the path is.

### 1.3.2 Path

In metric geometry, a path in a topological space X is a continuous map $f : I[a, b] \to X$. $I[a, b]$ denotes a sequence between a and b with small intervals.

In terms of our project, the term path is related to the learned manifold in the latent space . If $x_1$ and $x_2$ are two real frame images, and we feed them through the learned encoder, we will have two points $z_1$ and $z_2$ in the latent space respectively. The path is a curve on the manifold that connects these two points. In this paper, we want to calculate a shortest path on the manifold, i.e. the geodesic.

### 1.3.3 Geodesic

The shortest path in a manifold. It can be described by following formula:

$$\begin{aligned} \text{geodesic} &= \left\{\text{argmin}_{z \in \text{latent space}} \mathcal{L} : z\right\} \\ \mathcal{L} &= \sum_{n=0}^{N-1} \|g(z_n) - g(z_{n+1})\| \end{aligned} \tag{1.1}$$

, where z is point in latent space, and g is the decoder.

### 1.3.4 Route

The route is a sequence of points (i.e. images) in image space, and corresponds to a path in latent space.

A route is calculated by feeding each point in path through the decoder, resulting in a reconstructed image sequence.

$$\text{route} = \{g(z_i) : z_i \in \text{path}\} \tag{1.2}$$

In our experiments, we found that the generated images were often unclear, so to construct our route we select images from the training set that are most similar to the generated images. This is done by the squared Euclidean distance similarity metrics. The resulting image sequence is the route:

$$\text{route} = \{\text{argmin}_{x \in \text{dataset}} \|g(z) - x\| : z \in path\} \tag{1.3}$$

, where g is the decoder, and dataset is training images.

## 1.4 Organization of remaining chapters

In the Chapter 2, we describe the variational autoencoder.

In the Chapter 3, the mathematical foundation for VAE method developed by [Kingma and Welling, 2013] will be explained, giving the corresponding SGVB estimator and AEVB algorithm.

In the Chapter 4, we train the VAE model with images extracted from a touring video, and display the visualization of the learned latent space manifold, with a comparison with others' work.

In the Chapter 5, we explain our method for selecting a path on the learned manifold, and the method for producing the corresponding route in real environment, then finally the way we evaluate the results.

In the Chapter 6, we show three cases of navigation. We select a path for each, produce the corresponding route, and then evaluate the results.

In the Chapter 7, we analyze the reasons for non-ideal discontinuous routes, which can't be followed by robots with computer vision techniques in practice, and finally put forward some possible future work for improvement.

# Chapter 2

# Related work overview

There are discriminative models and generative models. Discriminative models model the dependence of unobserved (target) variables y on observed variables x[Wikipedia contributors, 2004]. Generative models model the data structure, sometimes together with latent variable structure, and usually in structured probability models in terms of graph and factors. The main difficulty of training generative models with latent variables is, given data X, to infer the "posterior" distribution over latent variables $p(z|x)$, i.e. what may have produced this data.

There can be various approaches to learn generative models: Variational Autoencoders attempt to learn log-probability of data using approximate inference. 2.1 and 2.2 will give a brief review for the approximate inference, and 2.3 will introduce Variational Autoencoders, one state-of-the-art method in approximate inference field. EM[Neal and Hinton, 1998] and Sparse Coding Models attempt to learn with an approximate posterior iteratively. This will not be discussed in this work. Generative Adversarial Networks use a different method for training that does not require making a posterior approximation. It will be mentioned in 2.4, and 2.4 will also show how can it be combined with VAE to improve the model power.

## 2.1 Inference in generative models

One problem that makes it difficult to train generative models is the problem of intractable inference. This problem arises because of the interactions between latent variables and visible variables. Here we give a brief introduction to the inference procedure in generative models, which is explained in detail in [Goodfellow et al., 2016]. Exact inference can be described as an optimization problem, approximate inference then can be described as approximating the underlying optimization problem.

Assume a probabilistic model consisting of observed variables X and latent vari-ables Z. Our aim is to find model parameters $\theta$ that maximize the log-probability of observed data: $\log p(\theta; X)$. For most non-trivial models, it is difficult to compute it because it is costly to marginalizing out Z. So instead, a lower bound $\mathcal{L}(\theta, q; X)$ on $\log p(\theta; X)$ is computed as Eq 2.1. This lower bound is called evidence lower bound (ELBO), or negative variational free energy.

$$\mathcal{L}(\theta, q; X) = \log p(\theta; X) - D_{KL}(q_\Phi(Z|X)||p_\theta(Z|X)) \tag{2.1}$$

The difference between $\log p(\theta; X)$ and $\mathcal{L}(\theta, q; X)$ is given by the KL divergence and between approximate the true posterior , and $\mathcal{L}(\theta, q; X)$ is at most as equal with $\log p(\theta; X)$, and for $q_\Phi(Z|X)$ that better approximations of $p_\theta(Z|X)$ , the lower bound $\mathcal{L}(\theta, q; X)$ will be closer to $\log p(\theta; X)$. Eq 2.1 can further be re-arranged by some simple algebra:

$$\mathcal{L}(\theta, q; X) = \log p(\theta; X) - D_{KL}(q_\Phi(Z|X)||p_\theta(Z|X)) \tag{2.2}$$

$$= \log p(\theta; X) - E_{q_\phi(Z|X)}[\log \frac{q_\phi(Z|X)}{p_\theta(Z|X)}] \tag{2.3}$$

$$= E_{q_\phi(Z|X)}[-\log q_\phi(Z|X) + \log p_\theta(Z, X)] \tag{2.4}$$

And inference is the procedure of finding the q that maximize $\mathcal{L}(\theta, q; X)$.

## 2.2 Approximation methods and variational inference

Compared with discriminative models for inference procedure, generative models face the problem of approximating the intractable probabilistic distribution, such as $\log p_\phi(Z|X)$ in Eq 2.4, or expectations with respect to this distribution. This is because dimensionality of the latent space is too high to sample densely or because the posterior distribution has a highly complex form for which expectations are not analytically tractable[Nasrabadi, 2007]. Two main methods for approximation are stochastic and deterministic approximations. Stochastic approximations are mainly based on mainly Markov chain Monte Carlo(MCMC) methods [Metropolis et al., 1953], where a Markov chain is constructed over the hidden variables whose stationary distribution is the posterior of interest . By repeatedly running the Markov chain, the sample converges to a fair sample from the true posterior distribution. The restricted Boltzmann machine (RBM) [Smolensky, 1986][Hinton et al., 2006] is one main successful application for this method. But sampling methods are computationally demanding, exhibiting high variance, and it can be difficult to know whether a sampling scheme is generating independent samples from the required distribution[Nasrabadi, 2007]. Deterministic approximations, on the other hand, are based on analytical approximation to the posterior distribution, including methods like variational inference and finite element method.

Variational inference restricts the optimization problem in 2.1 to a reduced set of more convenient distributions, usually a family of distributions. It then minimizes the Kullback-Leibler (KL) divergence from the variational distribution to the posterior distribution, finding the member in the family that is closest in KL divergence to posterior. This transforms complex inference probability problems into high-dimensional optimization problems[Jordan et al., 1999]. An common approach is to use a factorized form[Neal and Hinton, 1998][Hinton and Van Camp, 1993] of approximate distributions, which corresponds to an approximation framework developed in physics called mean field theory[Parisi, 1992]. Another approach is to restrict the family of approximating distributions to a parametric distribution $q(Z|\omega)$ family governed by a set of parameters $\omega$. This approach is known as structured or fixed-form Variational Bayes[Honkela et al., 2010].

Variational inference was first pioneered by Michael Jordan's lab[Saul et al., 1996][Jordan et al., 1999][Wainwright et al., 2008], while in parallel [Neal and Hinton, 1998] and [Hinton and Van Camp, 1993] carried the mean-field approach research for mixtures of experts. Here we are interested in Variational inference with continuous latent variables. [Hinton et al., 1995] introduce wake-sleep algorithm, in the "wake" phase, neurons are driven by recognition connections, and generative connections are adapted to increase the probability that they would reconstruct the correct activity vector in the layer below; in the "sleep" phase, neurons are driven by generative connections, and recognition connections are adapted to increase the probability that they would produce the correct activity vector in the layer above. But the drawback is that it requires a concurrent optimization of two objectives in both phases that would not corresponds to the optimization of marginal likelihood[Kingma and Welling, 2013], and it can only be able to train a inference network on latent variable values that are with high probability[Goodfellow et al., 2016].

[Hoffman et al., 2013] claimed that in practice, traditional variational inference can't be easily scale to the high-dimension and complex data problem and is solved with a coordinate ascent algorithm, iterating between re-analyzing every data point in the data set and re-estimating its hidden structure, and this is inefficient for large datasets. They derived a more efficient and scalable algorithm, i.e. stochastic variational inference, by using stochastic optimization[Robbins and Monro, 1951] for variational inference. It iterates between subsampling the data and adjusting the hidden structure based only on the subsample. [Paisley et al., 2012] introduced the use of control variates to reduce the variance of the stochastic search gradient mean-field when variational inference is a method for approximate Bayesian posterior inference. It approximates a full posterior distribution with a factorized set of distributions by maximizing a lower bound on the marginal likelihood. Although in the context of discriminative modelling, [Salimans et al., 2013] introduced a new interpretation of fixed-form variational approximation for learning the natural parameters of exponential-family distributions in the stochastic variational inference. Regarding directed probabilistic models with continuous latent variables, and the drawbacks of mean-field method that it requires analytical solutions of expectations, [Kingma and Welling, 2013] and [Rezende et al., 2014] introduced an alternative way by reparameterizing the variational lower bound, and yeilding a simple differentiable unbiased estimator called SGVB. This estimator can be used for efficient approximate posterior inference in almost any model with continuous latent variables.

## 2.3 VAE

[Kingma and Welling, 2013] and [Rezende et al., 2014] introduced a reparameterization trick for the variational lower bound in stochastic variational inference. In [Kingma and Welling, 2013] for the case of an i.i.d. dataset, continuous latent variables per data point, and a neural network for the recognition model, they named it variational auto-encoder. Variational autoencoders can also use convolutional networks in their encoders/decoders when dealing with image data. The algorithm for training a VAE will be explained in Chapter 3. Fig 2.1 shows a visualization of learned data manifold with two dimensional latent space, learned with VAE.

Variational Autoencoders have been applied to a host of tasks. [Kingma et al., 2014] combined semi-supervised learning with VAE. They performed nearest-neighbour and TSVM classification with features generated by the latent-feature discriminative model. It allows for effective generalization from small labelled data sets to large unlabelled ones. [Pu et al., 2016] also used VAE for semi-supervised learning in classifying image labelling and captions. But they used deep generative deconvolutional network (DGDN) as a decoder of the latent image features, and a deep Convolutional Neural Network(CNN) as an image encoder. [Kulkarni et al., 2015] used VAEto learns a representation of face and chair images, with disentangled and semantically interpretable latent variables. When given a single input image, the learned model can generate new images of the same object with variations in pose and lighting. [Nash and Williams, 2017] introduced shapeVAE, i.e. VAE equippped with hierarchical architecture, to automatically synthesize and analyze 3D objects. [Semeniuta et al., 2017] combined VAE with a hybrid architecture that blends fully feed-forward convolutional and deconvolutional components with a recurrent language model, and reported several attractive properties for text generation. [Yan et al., 2016] proposed disentangling CVAE (Conditional VAE) that can be used for image reconstruction and completion.

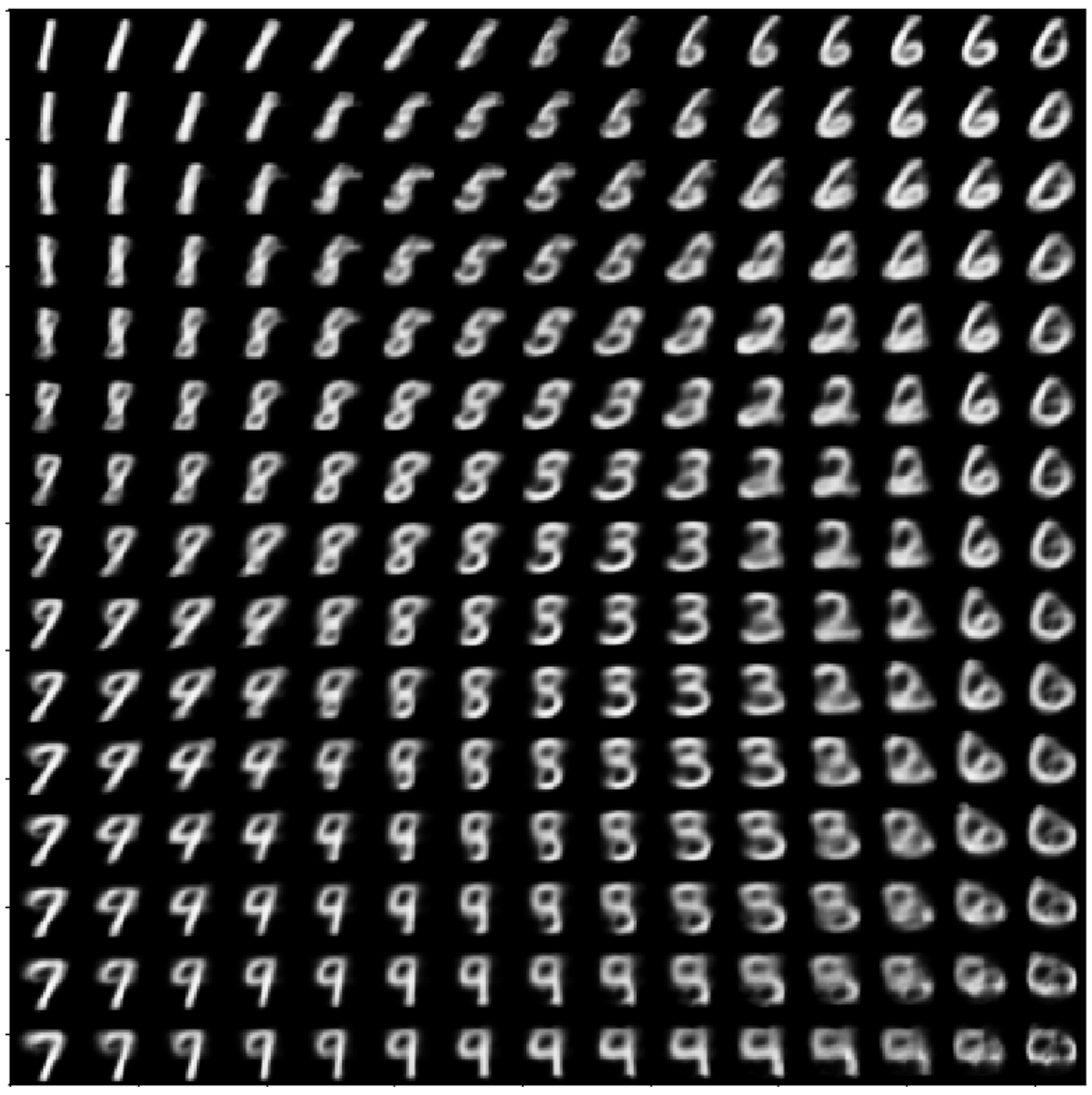

Figure 2.1: Visualization of learned MNIST data manifold. The outcome is the result of running the variational autoencoder with MLP code from: `https://github.com/keras-team/keras/blob/master/examples/variational_autoencoder.py`

## 2.4 Further development of VAE

VAEs often fail to generate clear images. [Chen et al., 2016] suggested that highly expressive inference models are essential in presence of a strong decoder to learn global representations that are useful for downstream tasks like classification. They combined VAE with neural autoregressive models such as RNN, MADE and Pixel-RNN/CNN, to improve the density estimation of VAE and make up the lack of right hierarchical structure of autoregressive models. [Kingma et al., 2016] demonstrated that VAE coupled with IAF, inverse autoregressive flow (IAF), is competitive with neural autoregressive models in terms of attained log-likelihood on natural images, while allowing significantly faster synthesis.

[Goodfellow et al., 2014] introduced generative adversarial networks (GANs), which are another generative modelling approach. And several works seek to fuse

VAE and GANs, trying to overcome the limitations of both.

GANs are based on a game theoretic scenario, the generator network competes against an adversary. The generator network tries to fool the adversary with the generated images, while the adversary attempts to distinguish between the real images and generated ones. But practically, training GANs is difficult and often leads to oscillatory behavior, divergence, or modeling only part of the data distribution, the phenomenon called mode collapse[Dosovitskiy and Brox, 2016] . [Larsen et al., 2015] added an adversarial loss to the variational evidence lower bound objective. They pointed out that element-wise measures such as the squared error are not suitable for image data, and used a feature-wise metric produced by GANs. [Dosovitskiy and Brox, 2016] also added the adversarial training to the VAE model. They claimed that the element-wise loss function tends to lead to over-smoothed results, and proposed a class of loss functions named perceptual similarity metrics with adversary training. [Mescheder et al., 2017] introduced an auxiliary discriminative network that allows to rephrase the maximum-likelihood problem as a two-player game, hence establishing a principled connection between VAEs and Generative Adversarial Networks (GANs). [Rosca et al., 2017] used the GANs to discriminate between real and model-generated data, with a reconstruction loss given by an auto-encoder. Specifically, they replaced the intractable likelihood by a synthetic likelihood, and replaced the unknown posterior distribution by an implicit distribution; and both synthetic likelihoods and implicit posterior distributions are learned using discriminators.

# Chapter 3

# Variational autoencoder

In this chapter, we will describe the variational autoencoder invented by [Kingma and Welling, 2013] and [Doersch, 2016]. They tackled with the intractability problem of generative models by adjusting the models to maximize the lower bound on the log-probability of the data. To use the famous back-propagation and gradient descent method in neural network, they reparameterized the lower bound.

## 3.1 Variational Autoencoder model introduction

Assuming that the data set $X = \{x_i\}^N$ consists of N i.i.d samples of some continuous variable x. x is generated by some random process, involving an unobserved continuous latent variable z. We want to learn the distribution of data set X, and its distribution $P(X|Z)$ in latent space. So a generative model is built, the marginal likelihood $\log p_\theta(X)$, it is a sum over the marginal likelihoods of individual datapoints:

$$\log p_\theta(X) = \sum_{i=1}^{N} \log p_\theta(x_i) \tag{3.1}$$

$$= \sum_{i=1}^{N} p_\theta(z) P_\theta(x|z) \tag{3.2}$$

Fig 3.1 is the graph representation of VAE, solid lines denote the generative model $p_\theta(z)p_\theta(x|z)$, dashed lines denote the recognition model, the variational approximation $q_\phi(z|x)$ to the intractable posterior $p_\theta(z|x)$. Neural networks are used as standard function approximators for both the encoder and the decoder. In this project, we use a convolution neural network.

But the problem here is that the true parameters $\theta$ and latent variables z are unknown to us. The solution is that we train a VAE by adjusting our model to maximize the lower bound on the log-probability of the data.

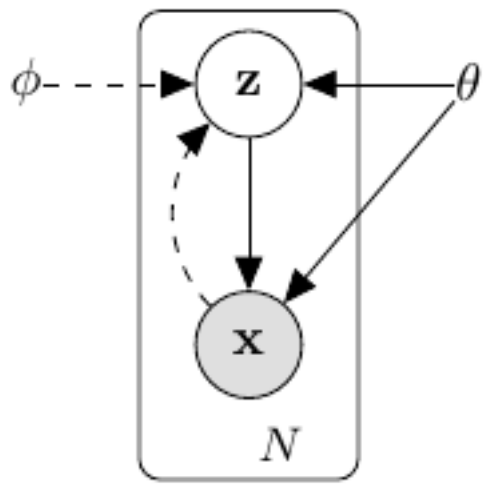

Figure 3.1: directed graphical model for Variational Autoencoder

Variational inference is performed on latent variables, and then maximum likelihood(ML) or maximum a posterior(MAP) inference is performed on global parameters, i.e. the two parameters $\phi$ and $\theta$ are optimized jointly. The main method to realize this is to drive a lower bound estimator for directed graphical models with continuous latent variables z and reparameterize z with a deterministic function of a white noise variable. This "reparameterization trick" allows us to propagate the gradient back through the network. Therefore, with the lower bound estimator and neural network structure, highly successful backpropagation with Gradient Descent and other deep learning strategies, like dropout, hyperparameter schemes,etc, can be used.

## 3.2 Variational lower bound

To calculate the Variational lower bound, further rewrite the marginal likelihoods of individual data point, i.e. $logp_\theta(x_i)$:

$$\log p_\theta(x_i) = E_{q_\phi(z|x_i)}[\log \frac{q_\phi(z|x_i)}{p_\theta(z|x_i)}] + \mathcal{L}(\theta, \phi; x_i) \tag{3.3}$$

$$= D_{KL}(q_\Phi(z|x_i)||p_\theta(z|x_i)) + \mathcal{L}(\theta, \phi; x_i) \tag{3.4}$$

This way, the latent parameter z is introduced. Here, $\mathcal{L}(\theta, \phi; x_i)$ is the lower bound on the marginal likelihood of datapoint $x_i$, as the KL-divergence is non-negative. And the maximum of the lower bound occurs when KL divergence vanishes, i.e. $q_\Phi(z|x_i) = p_\theta(z|x_i)$, the approximate distribution exactly the same as the true distribution. Thus, maximizing the marginal likelihood $logp_\theta(X)$, is equivalent to maximizing the lower bound $\mathcal{L}(\theta, \phi; x_i)$, plus an error term corresponding to the KL divergence $D_{KL}(q_\Phi(z|x_i)||p_\theta(z|x_i))$.

$\mathcal{L}(\theta, \phi; x_i)$ can be deducted as following:

$$\mathcal{L}(\theta, \phi; x_i) = -E_{q_\phi(z|x_i)}[\log \frac{q_\phi(z|x_i)}{p_\theta(z|x_i)}] + \log p_\theta(x_i) \tag{3.5}$$

$$= E_{q_\phi(z|x_i)}[-\log q_\phi(z|x_i) + \log p_\theta(z|x_i)] + \log p_\theta(x_i) \tag{3.6}$$

$$= E_{q_\phi(z|x_i)}[-\log q_\phi(z|x_i) + \log p_\theta(z, x_i)] \tag{3.7}$$

$$= -D_{KL}(q_\phi(z|x_i))||p_\theta(z)) + E_{q_\phi(z|x_i)}[\log p_\theta(x_i|z)] \tag{3.8}$$

Then in order to maximize $\log p_\theta(X)$ for the training dataset under the generative process, $\mathcal{L}(\theta, \phi; x_i)$ should be differentiated and optimized w.r.t $\phi$ and $\theta$. So we get the loss function for the model, that is $-\mathcal{L}(\theta, \phi; x_i)$.

## 3.3 SGVB estimator and AEVB algorithm

### 3.3.1 Generic SGVB estimator

In order to solve the problem mentioned in last part, variational inference is applied by restricting the range of distributions of variables to a family, such that $D_{KL}(q_\Phi(z|x_i)||p_\theta(z|x_i))$ is minimized or equivalently choosing a distribution w.r.t q that has most capacity, and then seeking the member in family for which the lower bound is maximized.

There are two ways to infer a posterior for parameters w.r.t the lower bound: approximate the posterior of z in the form of $q_\phi(z|x)$; and the fully variational Bayesian method for inferring a posterior over all the parameters. This project uses the former.

What's more, this approximation involves **Reparameterization of the Variational Lower Bound**, which results in a simple Monte Carlo estimator of $\mathcal{L}(\theta, \phi; x_i)$ that is differentiable and unbiased. This resulting estimator is Stochastic Gradient Variational Bayes estimator. This trick is necessary because Stochastic Gradient Descent method is applied in neural network as back-propagation, which can only handle stochastic inputs but not stochastic units that sample z from x.[Doersch, 2016]

[Kingma and Welling, 2013] proved that it is often possible to express z as a deterministic and differentiable transformation: z=$g_\phi(\epsilon, x)$, where $\epsilon$ is an auxiliary/noise variable with independent distribution $p(\epsilon)$, and $g_\phi(.)$ is some vector-valued function parameterized by $\phi$:

$$\tilde{z} = g_\phi(\epsilon, x), \text{with } \epsilon \sim p(\epsilon) \tag{3.9}$$

They also provided some methods for choosing such $p(\epsilon)$ and $g_\phi$ w.r.t different family categories chosen for distribution of $q_\phi(z|x)$.

Thus, apply this reparameterization trick on the variational lower bound: for

$$\tilde{z}_{i,l} = g_\phi(\epsilon_{i,l}, x_i), \text{with } \epsilon_l \sim p(\epsilon) \tag{3.10}$$

, expression of the lower bound ( version Eq 3.7 ) can be rewritten as following:

$$\tilde{\mathcal{L}}(\theta, \phi; x_i) = \frac{1}{L} \sum_{l=1}^{L} [\log p_\theta(z_{i,l}, x_i) - \log q_\phi(z_{i,l}|x_i)] \tag{3.11}$$

(Version Eq 3.8) can be rewritten as following.

$$\tilde{\mathcal{L}}(\theta, \phi; x_i) = -D_{KL}(q_\phi(z|x_i))||p_\theta(z)) + \frac{1}{L} \sum_{l=1}^{L} [\log p_\theta(x_i|z_{i,l})] \tag{3.12}$$

The latter is used when the $D_{KL}$ term can be solved analytically.

### 3.3.2 Generic AEVB algorithm

As the SGVB estimator is constructed, the algorithm to update parameters are as following:

initialization of $\theta, \phi$;

**while** *loss not converges* **do**

1. Randomly draw a mini-batch of M data points;
2. Randomly sample the auxiliary variable $\epsilon$ from $p(\epsilon)$;
3. Calculate the gradients of $\bigtriangledown_{\theta,\phi}\tilde{\mathcal{L}}(\theta, \phi; x_i)$
4. Update $\theta, \phi$ using some gradients descent method.

**end**

**Algorithm 1:** Generic AEVB algorithm

### 3.3.3 SGVB estimator and AEVB algorithm for variational autoencoder

This part will apply the above explained AEVB algorithm to Variational Autoencoder.

Let the prior over the latent variable z the centered isotropic multivariate Gaussian, i.e. $p_\theta(z) = \mathcal{N}(z, 0, I)$. And let the variational posterior $p_\theta(x|z)$ be a multivariate Gaussian:

$$\log q_\phi(z|x_i) = \log \mathcal{N}(z; u_i, \sigma_i^2 I) \tag{3.13}$$

Note that the true posterior $p_\theta(z|x)$ is usually intractable in practice, so it is modelled as an approximate Gaussian form with an approximate diagonal covariance.

SGVB estimator for the Variational Autoencoder model is as following (Eq 3.8):

$$\tilde{\mathcal{L}}(\theta, \phi; x_i) = -D_{KL}(q_\phi(z|x_i))||p_\theta(z)) + \frac{1}{L}\sum_{l=1}^{L}[\log p_\theta(x_i|z_{i,l})] \tag{3.14}$$

, where the KL term that can be calculated analytically, and the decoding term $\log p_\theta(x_i|z_{i,l})$ is a Gaussian neural network. Suppose z has J dimensions, and and $\mu_j$, $\sigma_j$ denote the j-th element for the parameters $\mu$ and $\sigma$ respectively.

$$-D_{KL}(q_\phi(z|x_i))||p_\theta(z)) = E_{q_\phi(z|x_i)}[\log \frac{q_\phi(z|x_i)}{p_\theta(z)}] \tag{3.15}$$

$$= E_{q_\phi(z|x_i)}[\log q_\phi(z|x_i) - \log p_\theta(z)] \tag{3.16}$$

$$= \int \mathcal{N}(z; \mu_i, \sigma_i^2) \log \mathcal{N}(z; \mu_i, \sigma_i^2) dz - \int \mathcal{N}(z; \mu_i, \sigma_i^2) \log \mathcal{N}(z; 0, I) dz \tag{3.17}$$

$$= -\frac{J}{2}\log(2\pi) - \frac{1}{2}\sum_{j=1}^{J}(\mu_j^2 + \sigma_j^2) - \frac{J}{2}\log(2\pi) - \frac{1}{2}\sum_{j=1}^{J}(1 + log\sigma_j^2) \tag{3.18}$$

$$= \frac{1}{2}\sum_{j=1}^{J}(1 + (\sigma_{j,i)})^2 - (\mu_{j,i})^2 - (\sigma_{j,i})^2) \tag{3.19}$$

Thus, the resulting SGVB estimator is:

$$\tilde{\mathcal{L}}(\theta, \phi; x_i) = \frac{1}{2}\sum_{j=1}^{J}(1 + \log(\sigma_{j,i)})^2 - (\mu_{j,i})^2 - (\sigma_{j,i})^2) + \frac{1}{L}\sum_{l=1}^{L}[\log p_\theta(x_i|z_{i,l})] \tag{3.20}$$

Lastly, we make a summarize for the AEVB algorithm for the Variational Au-

toencoder:

Randomly initialize parameters $\theta, \phi$;

**while** *loss not converges* **do**

1. Randomly draw a mini-batch of M data points;
2. sample from the posterior $z_{i,l} \sim q_\phi(z|x_i)$ using the following reparameterization trick:

$$z_{i,l} = g_\phi(x_i, \epsilon_l) = \mu_i + \sigma_i \odot \epsilon_l, \text{and } \epsilon_l \sim \mathcal{N}(0, I)^a \tag{3.21}$$

3. calculate the gradient of the variational lower bound Eq 3.20;
4. update the parameter $\phi$ and $\theta$ using gradient descent method.

**end**

**Algorithm 2:** AEVB algorithm for the Variational Autoencoder

[a] $\odot$ is the element wise product

# Chapter 4

# Experiment:learning the map

## 4.1 Training data preparation

As explained in the Chapter 1.1 we should tour through a house, making a video of the tour for training the model. We use a video fragment from YouTube[1], staring from 0:37 to 3:57, which satisfies the requirement. We make the dataset set of 1000 images from this video, drawn with equal time intervals. The drawn images have the size of 720*1280*3. For efficiency of the training all images are re-sized to 60*60*3 before feeding in.

## 4.2 Training the VAE model

A convolutional and deconvolutional neural networks are used for the encoder and the decoder. The model structure is shown in Fig A.1. The detail parameter settings for the layers of the neural network is shown in Fig A. The structure of the neural network in this project drives much inspiration from `https://github.com/keras-team/keras/blob/master/examples/variational_autoencoder_deconv.py` accessed Nov.2017, and this project uses Keras for training Variational Autoencoder model. But the main difference here is the dimensionality of the latent space, we chose 4D instead of 2D. As the image dataset in this project is very complex, dimensionality of 2 is too low, Fig B.2 shows the manifold of 2D latent space model, it is an ugly learning result. A intuitive explanation for this is that, representation learning tries to recover or disentangle underlying factors for the data structure[Bengio et al., 2013], so each dimension in 4 D latent variable captures the four directions the camera goes, i.e. (move forward/backward), (move left/right), (tilt left/right),

[1]https://www.youtube.com/watch?v=jdNiWiXiJQ4&t=200s

(tilt up/down). In principle, there are 6 (x,y,z, roll,pitch,yaw), but usually we don't (tilt clockwise/counterclockwise) or (move up/down) when walking around a house.

For training the network, the adaptive learning rate method 'RMSprop' is adopted with Gradient descent. Batch size is 20, and epoch is 2000. Data is shuffled when feeding into the model. The loss, which is the negative lower bound on the log-probability of the training data under the VAE mode, i.e. the negative of Eq (3.21), is recorded, it shows that in Fig 4.1 with 2000 iterations, the loss is asymptotic to its local minimum. The code for building the model and running the experiments is at `https://github.com/augustkx/VAE_learning-a-representation-for-navigation`.

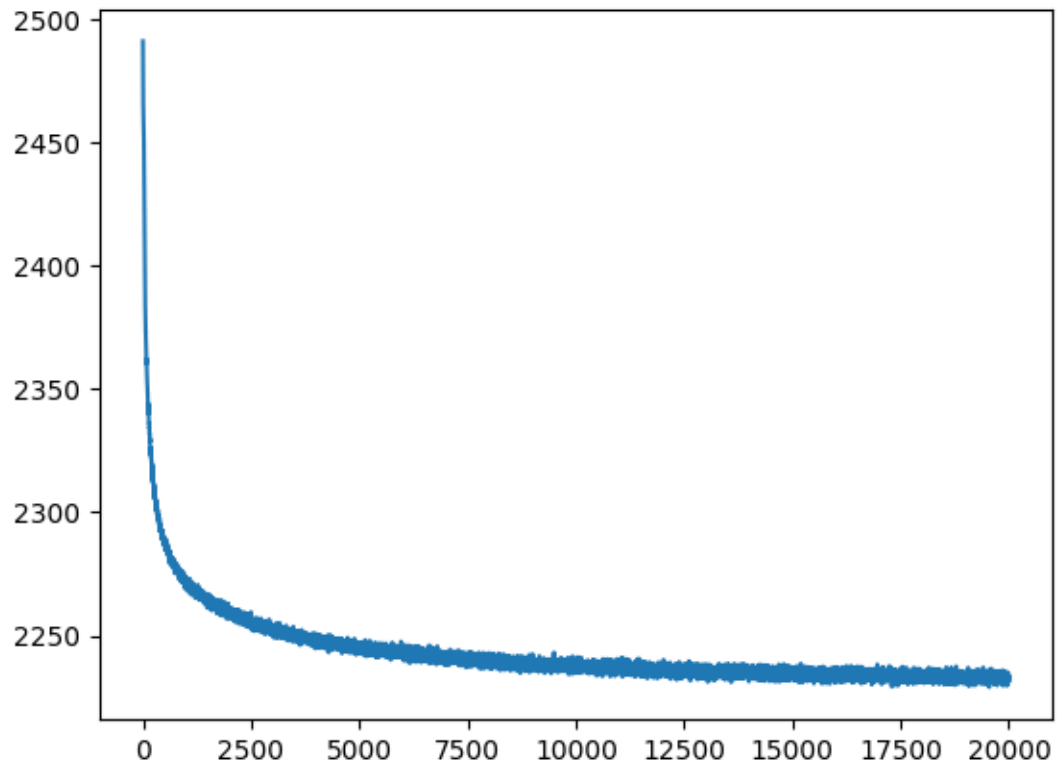

Figure 4.1: The loss value through the training. The loss function in a Variational Autoencoder is the negative lower bound of the log probability of the training data.

## 4.3 Map in latent space

### 4.3.1 Visualization

A slice of the latent space manifold is shown in Fig 4.2. It is produced by sampling a 2D grid of points in the 4 dimension latent space, and then feeding the points through the trained decoder, resulting in corresponding generated image for each point. When sampling, we set the last two dimensions to be linearly spaced between 0.05 and 0.95, and set values for remaining dimensions to 0. So the result is a 2-D grid, in terms of the last two dimensions, linearly spaced on the square bounded by 0.05 and 0.9. It shows that in this region of the latent space, the generated images can roughly be classified into 6 different scenes corresponding to the training images. For each of these 6 different scenes, one corresponding representative generated image is circled[2]. These 6 images are from nearby rooms, but the scenes in them change sharply. Between these 6 circled images, there are some images almost the same as of one them but more fuzzy, there are also filled with some blurry images that are the a blend of two or three or several typical images around them.

In order to have more intuitive understanding of the manifold, in Appendix B, more slices of manifold are presented.

[2]This is done manually and qualitatively, and you can also select some other images for representing the classification.

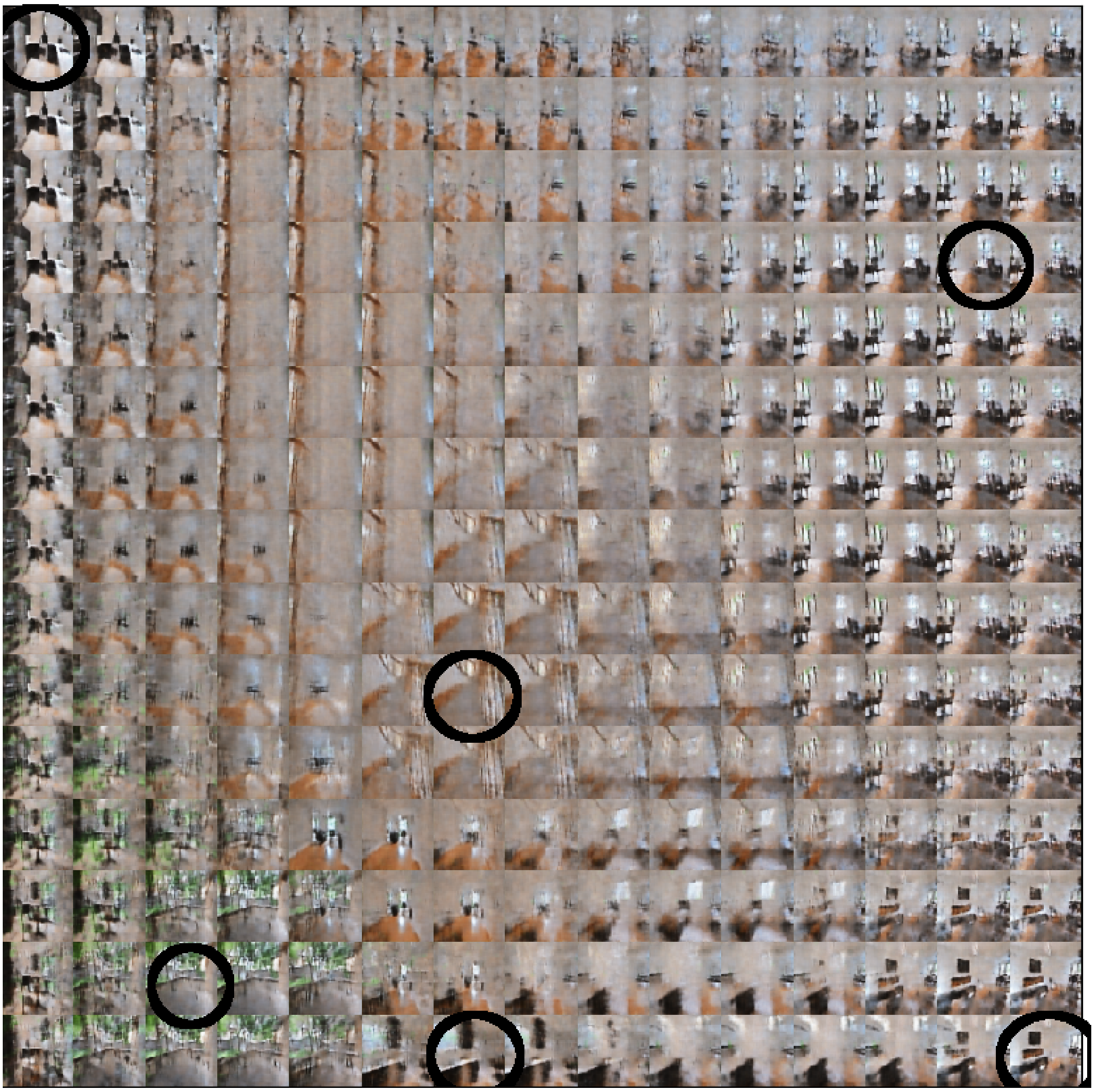

Figure 4.2: A slice of the latent space manifold with a qualitative and manual classification of the generated images marked by circles. This is to give a rough glimpse to the latent variable data structure on learned manifold.
To sample latent variable z, we set the last two dimensions to be linearly spaced between 0.05 and 0.95, and set values for remaining dimensions to 0. So the result is a 2-D grid, in terms of the last two dimensions, linearly spaced on the square bounded by 0.05 and 0.9.

### 4.3.2 Comparison with related work's outcome

[Doersch, 2016] produced samples from a VAE trained on MNIST dataset(shown in Fig 4.3), and reported while most of the digits look quite realistic, a significant number of them are in-between different digits. [Dosovitskiy and Brox, 2016] pointed out that VAE alone (Fig 4.4 left) tends to generate blurry images with complex image data. So the learning outcome of the VAE in this work is similar with other works.

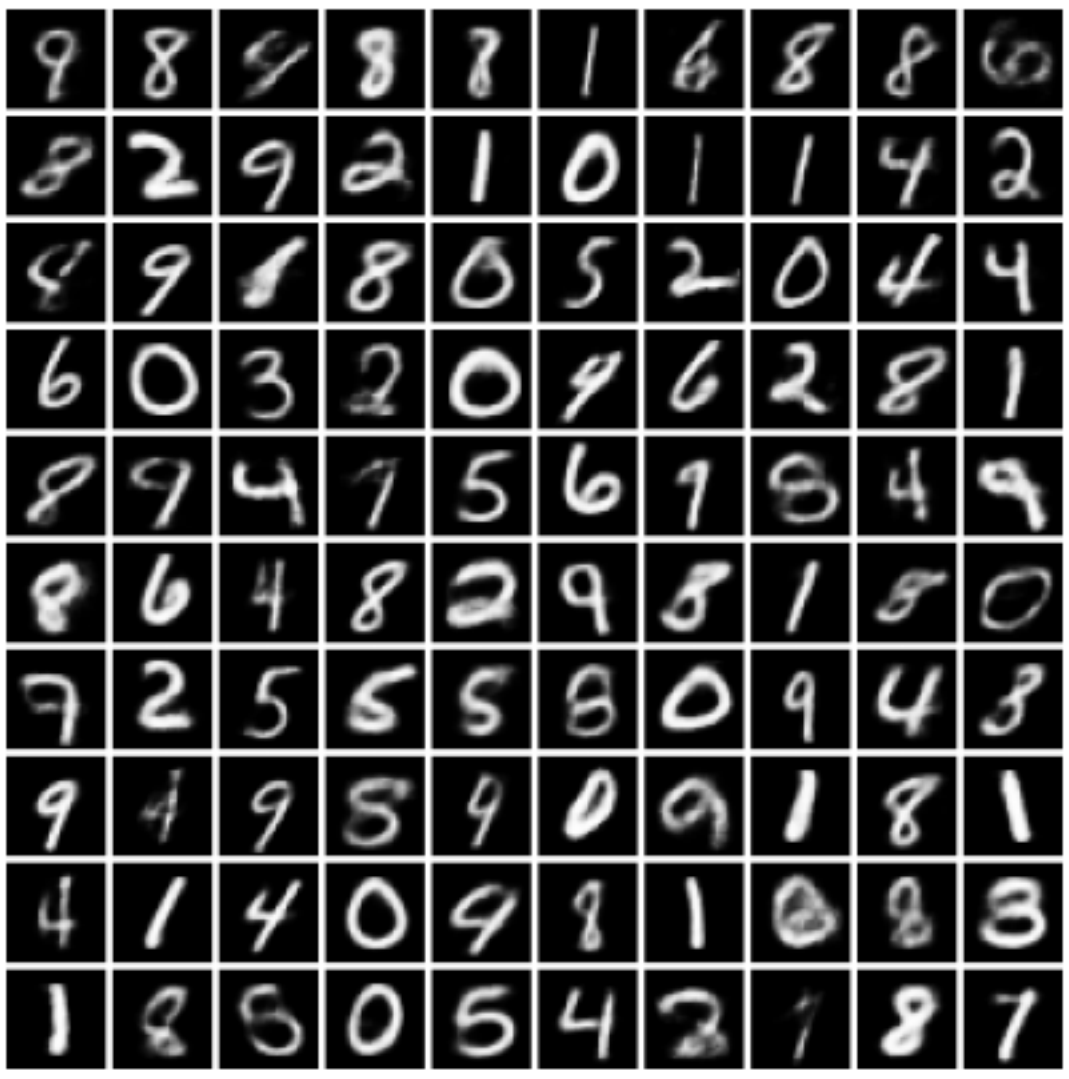

Figure 4.3: Sampling from a VAE trained on MNIST by [Doersch, 2016]. It can be seen that while most of the digits look quite realistic, a significant number are 'in-between' different digits. For example, the seventh digit from the top in the leftmost column is clearly in-between a 7 and a 9.

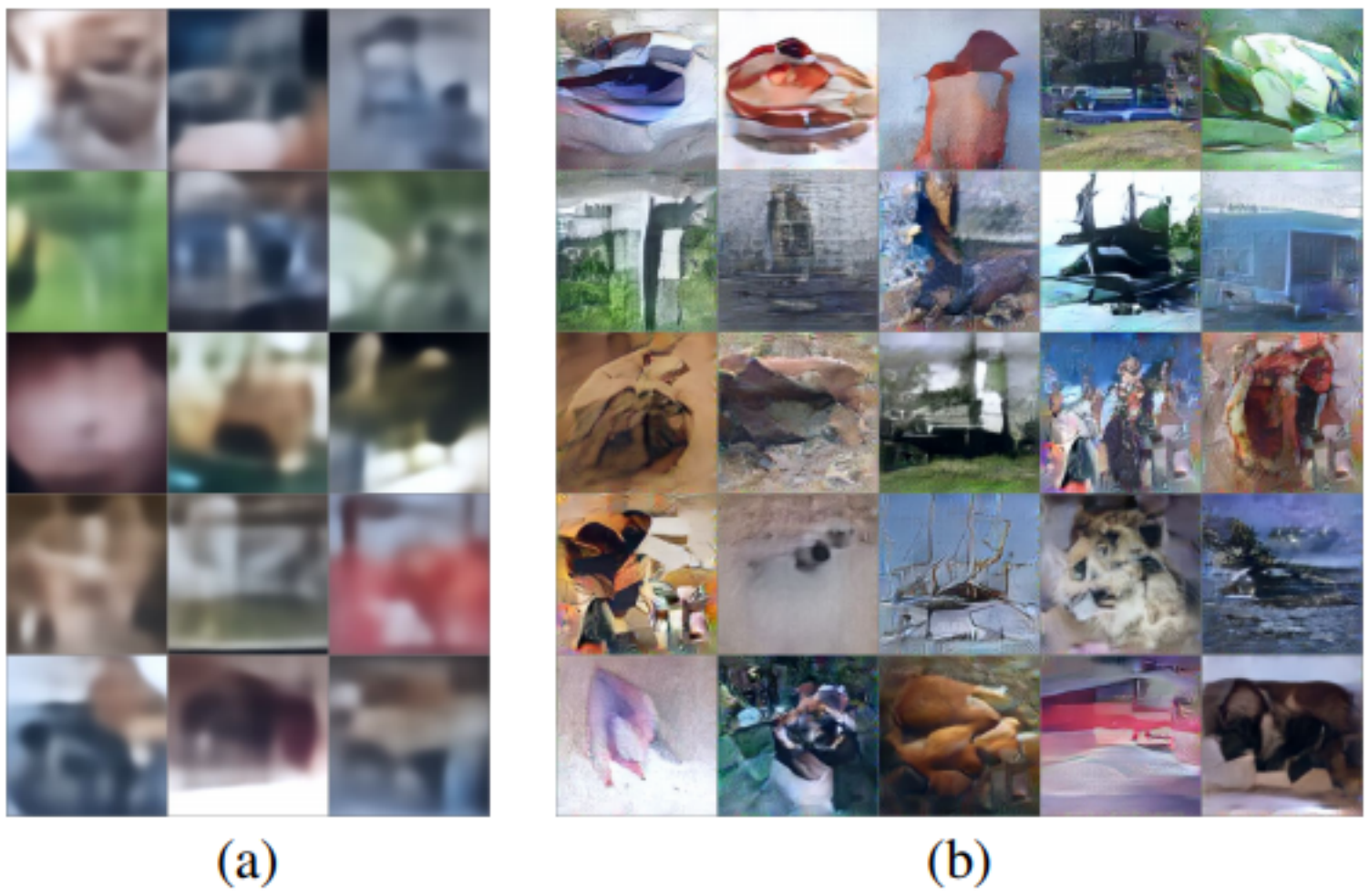

Figure 4.4: Reconstructions from VAE with the squared Euclidean loss (left) and the proposed DeePSiM loss (right) by [Dosovitskiy and Brox, 2016]

# Chapter 5

# Path planning method

In this chapter, we will illustrate two assumptions related to the computer vision techniques, and then make a hypothesis on the learned manifold. To test this hypothesis., 5.3 explains the method of selecting paths on the learned map, and producing corresponding routes, 5.4 describes the metrics for evaluating. And in the following chapter, we will use three cases to test the hypothesis, and evaluate the results.

## 5.1 Assumptions

We make two assumptions, which are related to to the technical problems with real robots:

1. In terms of navigation, the distance between each point should be short enough so the robot can follow the route, and move from the starting scene to destination step by step using assumed computer vision techniques pointed out in Chapter 1.1. By the computer vision techniques, we mean that given a video of the route to follow, the robot is able to follow this route in a real environment.

2. The starting and ending images are already on the toured line, i.e. in the video used to draw training data set. This assumption makes sure that the robot is able to lead itself through the path through computer vision techniques. If this assumption is not met, then the robot roams round until it finds itself in a location that has been toured when training.

## 5.2 Hypothesis

We make following hypothesis such that we select geodesic as the path.

The learned representation map is continuous and complete as Chapter 1.1 describes: continuity means that the neighbour points on the manifold represents real frames that also are neighbourhood geographically; completeness means that similar real frames will have different representations in the latent space, so each real frames can be recognized and distinguished by the model.

The generated images in the manifold in Fig 4.2, Fig C.1 and Fig C.2 indicate that the generated images on a slice are continuous from appearance, but as some of them are very blur, it can't be figured out easily if they correspond to real frames that are very near geographically. If this hypothesis is true, then our method for selecting a path in the latent space will have a corresponding continuous and reasonable route.

## 5.3 Navigation experiment design

### 5.3.1 Map the starting scene and destination to learned map

1. Randomly select two images from the training set as the staring and ending scenes. There are 1000 images in the training set, all of which are labeled from 1 to 1000 in a sequence of touring order. For real robot simulations, we should randomly shot two completely new images anywhere in the house as the starting and ending scenes. But for simplification, this project just selects in the training collection.

2. Feed the two images through the learned encoder to get the corresponding points in the Z latent space. We get $z_s$ and $z_d$.

3. Figure out a path in the latent space that connects the two points, such that the robot can follow the corresponding route in reality that lead the robot from the starting scene to the destination. As pointed out in Chapter 1.1, we choose geodesic (definition in Chapter 1.3), which is the shortest path in a manifold, as the path. The formula for geodesic is:

$$\begin{aligned} \text{geodesic} &= \left\{ \text{argmin}_{z \in \text{latent space}} \mathcal{L} : z \right\} \\ \mathcal{L} &= \sum_{n=0}^{N-1} \| g(z_n) - g(z_{n+1}) \| \end{aligned} \tag{5.1}$$

, where z is point in latent space, and g is the decoder.

This is because we have made the hypothesis that learned representation map is continuous and complete, so any path on the manifold is continuous, and regarding navigation the shortest the better.

### 5.3.2 Path planning with gradient descent method

There are four methods according to [REDFIELD, 2007] to compute a geodesic between two points on a manifold: 1. Using the midpoints of an approximate path between them; 2. Gradient descent method to iteratively update the path approximating the geodesic; 3. Numerically solving the system of differential equations; 4. Graph search methods in a finite graph.

In order to minimize $\mathcal{L}$ in Eq 5.1, this project will use the **gradient descent method** mentioned in [REDFIELD, 2007] to iteratively update the path approximating the geodesic on the latent space manifold. [1].

First, we initialize a path sequence. We have the straight line connect $z_s$ and $z_d$ in the latent space, and project it to the learned manifold. We use this curve for updating. The number of points along the route and path should be chosen as big enough to form a continuous route, while as small as possible to minimize the computing cost. In this project, 50 images are large enough to form a continuous path. So we set the path consisting of 50 points, which are chosen with equal intervals between each other along the initial straight line connecting $z_s$ and $z_d$.

Second, for a path consisting of several points, Gradient Descent method considers three consecutive points at a time and update the middle point of each of these sequences $z_i$ ,$z_{i+1}$ , $z_{i+2}$ of the path as it goes from one end of the path to the other.Specifically, this method first fixes $z_i$ and $z_{i+2}$ ,then selects a new middle point z from a neighborhood of the original $z_{i+1}$ which allows the path to better approximate the geodesic connecting $z_i$ and $z_{i+2}$. This is updated by this formula for $i \in [2, 3, ...N-1]$ in order and repeatedly:

$$\mathbf{min}\ \mathcal{L}(z) = \|g(z_i) - g(z)\| + \|g(z) - g(z_{i+2})\| \tag{5.2}$$

To minimize above formula, Gradient Descent method is used, in which $z_{i+1}$ is updated iteratively in the direction opposing the gradient $\bigtriangledown L$:

$$z_{i+1} \leftarrow z_{i+1} - \alpha \bigtriangledown LB : i+1 \in [2, 3, ...N-1] \tag{5.3}$$

, where N is the number of points between (and including) $z_s$ and $z_d$.

[1]It should be pointed out that the geodesic calculated using this method is not necessarily the global shortest path between two points[REDFIELD, 2007]. But this doesn't harm our experiment that much as long as we are more interested in whether the produced route is continuous or not.

### 5.3.3 Route producing and testing

As long as we get the updated sequence of points in latent space, we can get the corresponding route by feeding each through decoder: route $= \{g(z_i) : z_i \in path\}$. But as the VAE tends to generate many blurry images, as those in Fig 4.2, the generated route can't be used for navigation directly. So as pointed out in Chapter 1.3.4 we match each reconstructed image with a real frame in training collection, which is the most similar image. This is done by the squared Euclidean distance similarity metric. The resulting image sequence is the route:

$$\text{route} = \{\text{argmin}_{x \in \text{dataset}} \|g(z) - x\| : z \in path\} \tag{5.4}$$

## 5.4 Evaluate

We can roughly test the success of navigation by the continuity of the route. As the comparison base, we[2] manually select a route that is perfectly continuous in the environment. We denote the geodesic as $\text{path}_{\text{geodesic}}$, and the corresponding route as $\text{route}_{\text{geodesic}}$; And denote the path corresponds to the manually selected route as $\text{path}_{\text{manually}}$, and the manually selected route as $\text{route}_{\text{manually}}$. The $\text{route}_{\text{geodesic}}$ is calculated by Eq 5.4, the $\text{path}_{\text{manually}}$ is calculated by:

$$\text{path} = \{\text{encoder}(x) : x \in \text{route}\} \tag{5.5}$$

First, we count the number of image categories in $\text{route}_{\text{geodesic}}$ over that in $\text{route}_{\text{manually}}$. This indicates how good the produced route is.

Second, we feed the the ideal route selected manually through the trained encoder and decoder, resulting in the reconstructed route w.r.t. the manually selected route. If all the reconstructed images are unique and continuous, then there exists such a path in the latent space that has a corresponding continuous route in reality. In order to evaluate this ideal route, we calculate the distance between each pair of neighbour points in $\text{path}_{\text{manually}}$, and the distance between each point in $\text{path}_{\text{manually}}$ and a latent space representation of randomly selected image in the training set. If the former distanceis significantly much smaller than the second distance, we can say that the learned map captures the layout of the house well. This evaluates how good the learned manifold is.

Third, we evaluate the neighbour-image-distance distribution difference between $\text{route}_{\text{geodesic}}$ and $\text{route}_{\text{manually}}$, also the neighbour-point-distance distribution differ-ence between $\text{path}_{\text{geodesic}}$ and $\text{path}_{\text{manually}}$. This helps to figure out how to adjust the optimization function to improve the produced route.

[2]Human, as opposed to robot.

# Chapter 6

# Experiment: path planning

The layout of the house structure on the ground floor is shown in Fig 6.1. Three navigation cases will be about navigation tasks between different rooms.

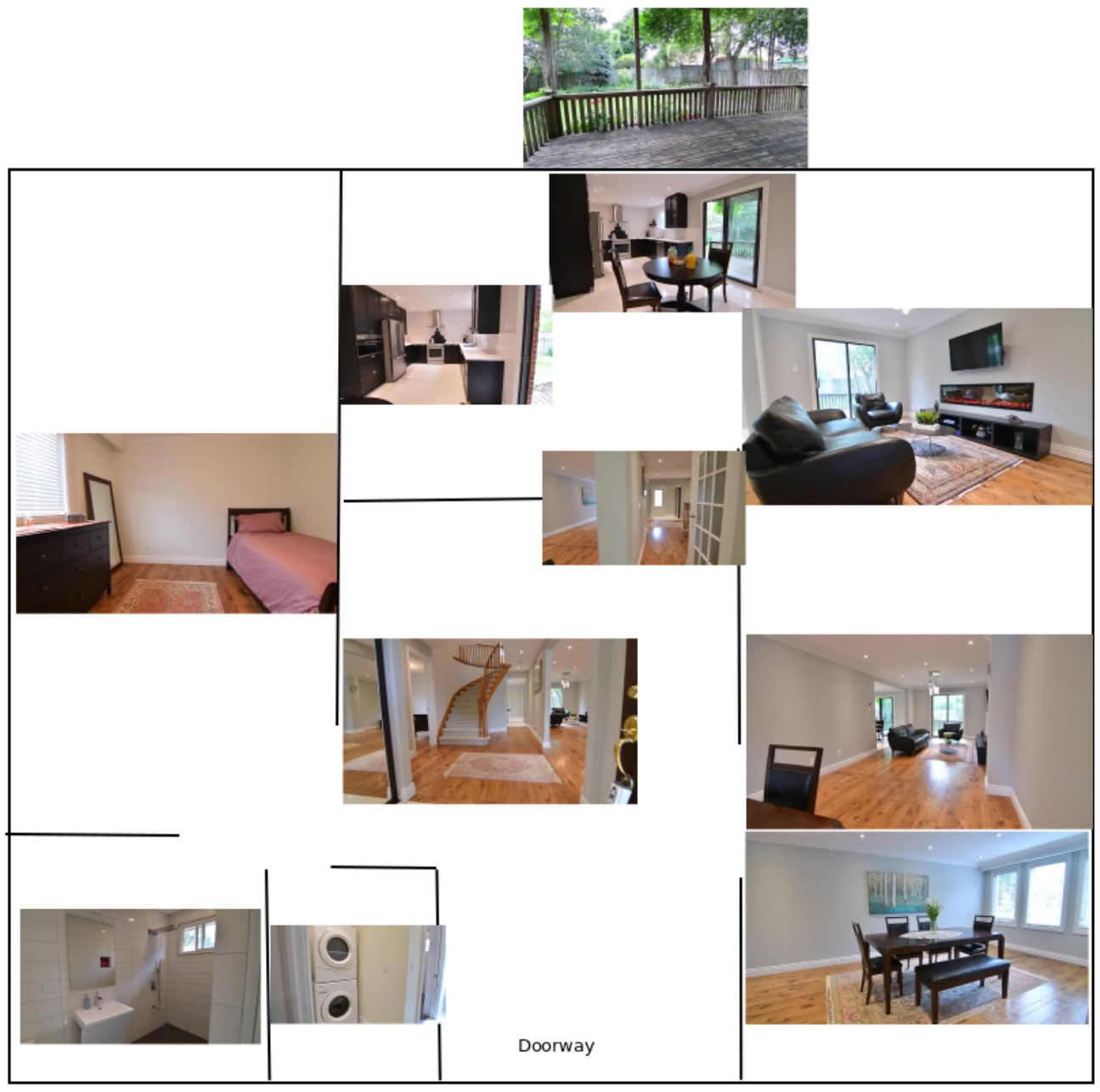

Figure 6.1: The layout of the house structure on the ground floor. Three navigation cases will be about navigation task between different rooms.

## 6.1 Case one

**Starting scene: the first image, the doorway; Destination: the 300-th image, the kitchen.**

### 6.1.1 Route producing

With procedures in Chapter 5.3, we select the geodesic described in Chapter 1.3 as the path. We initialize a curve and update it with Gradient Descent method. Fig 6.2 top shows the generated images from the resulting route. Fig 6.2 bottom is a route of real frames which are the most similar ones with the generated images, as is explained in 5.3.3 and 1.3.4. One thing worth noting is that the initialized curve is almost the same as the updated geodesic, indicating some properties of the learned manifold.

The route consists of four scenes, through doorway to dinning room, and then to the kitchen. The route is reasonable and short in the sense that we actually have to pass these four scenes. But the route is discontinuous, and abruptly jumps between rooms. It's clear that it could not be followed by a route-following robot.

As our model can't figure out a method to get a successful route, we try to figure out if there exists such a path in the latent space that has a corresponding continuous route in reality. Fig 6.3 is the ideal route selected manually, and Fig 6.4 is the reconstructed route that produced by feeding the images of the manually selections into the trained encoder and decoder. Fig 6.4 shows that for each frame along the ideal route there is a corresponding unique latent variable that represents it. But as the generated images are quite different from the route in Fig 6.2 top, it indicates that the ideal path in the latent space is not a geodesic.

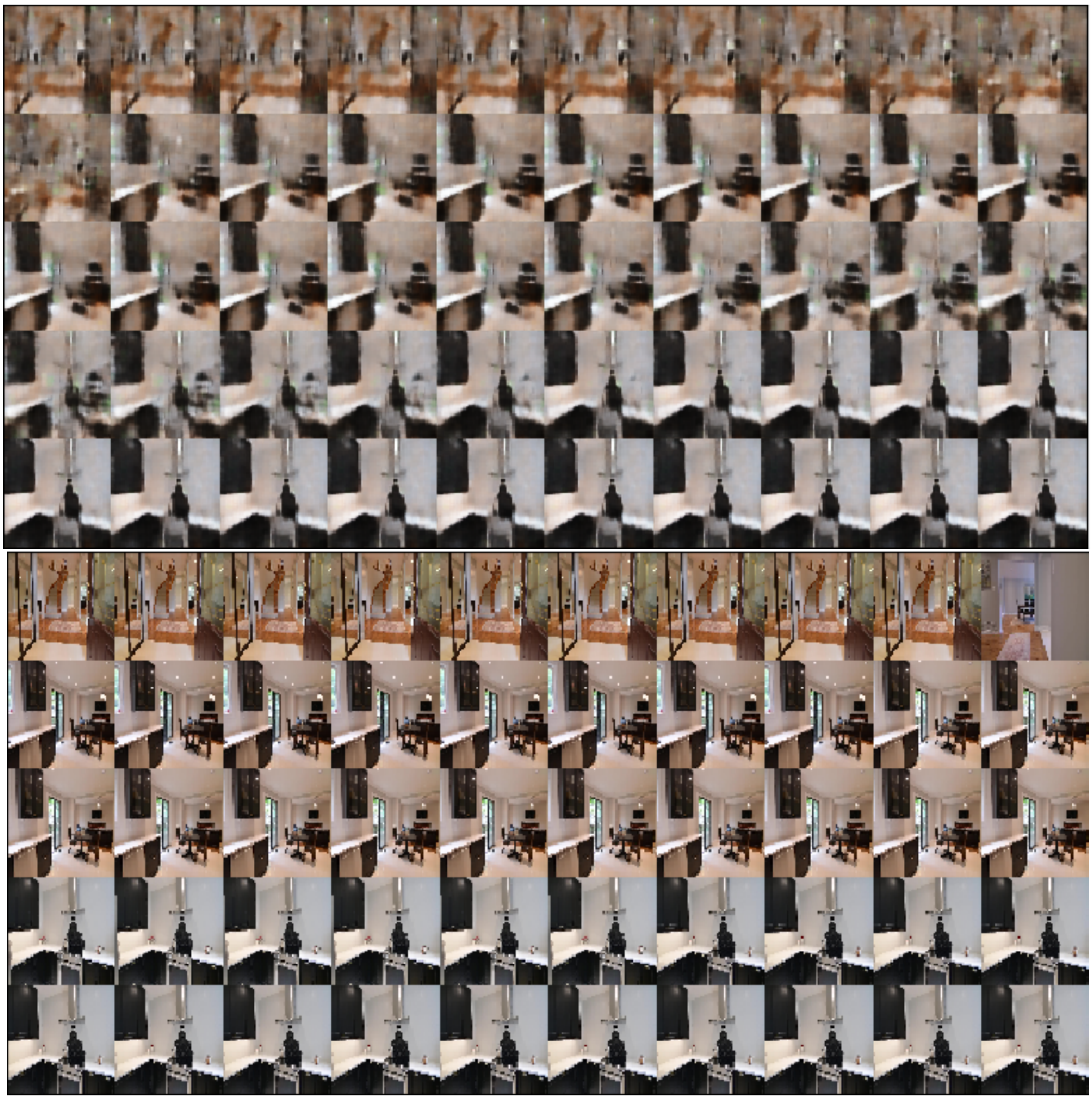

Figure 6.2: Experiment Case one: visualization of the route. The procedure for selecting such a path is explained in detail in Chapter 5. Top: route of generated images; Bottom: route of real frames, each image is the most similar real frames with the corresponding generated one, by the squared Euclidean distance similarity metric.

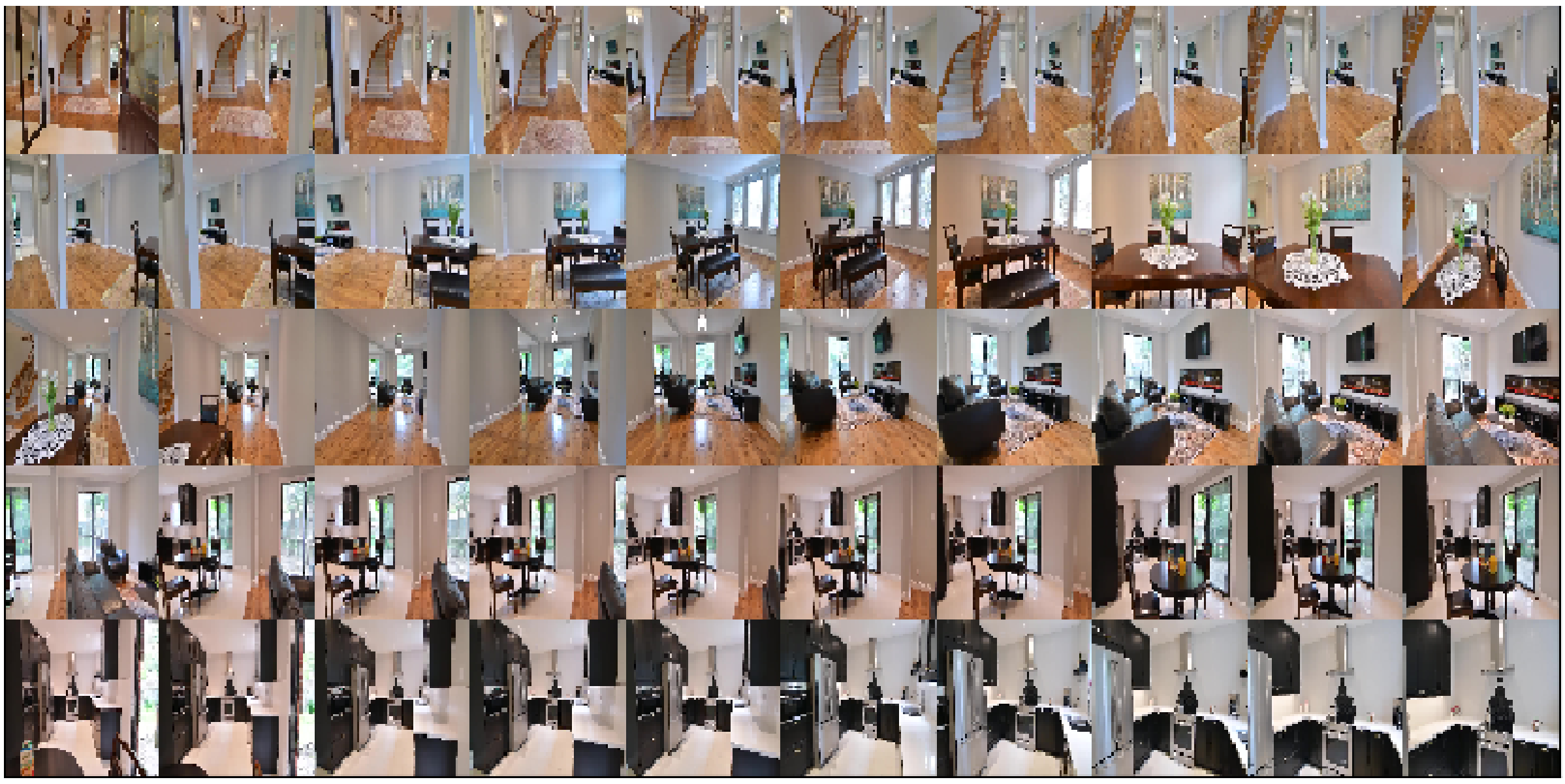

Figure 6.3: Experiment Case one: the ideal route selected manually. In this route, each neighbouring image pair is continuous geographically.

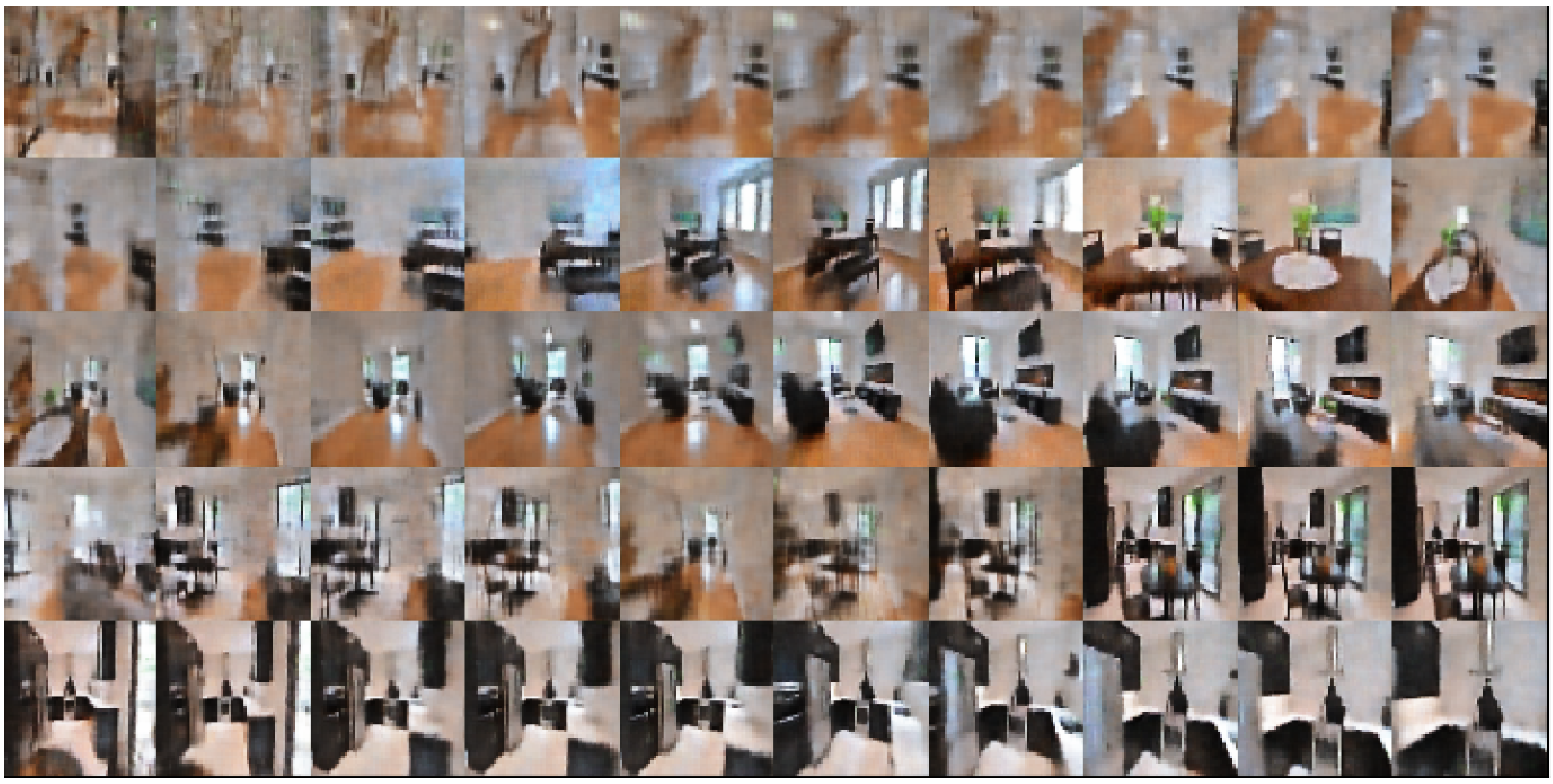

Figure 6.4: Experiment Case one: reconstructed route from an ideal route selected manually. It is produced by feeding each image along the route in Fig 6.3 through encoder and decoder.

### 6.1.2 Evaluation

First, the number of image categories in $\text{route}_{\text{geodesic}}$ over that in $\text{route}_{\text{manually}}$ is 4/50, as indicated by Fig 6.2 Bottom and Fig 6.3.

Second, we evaluate the manually selected route. Fig 6.5 Left is the distribution of difference of two distances, i.e. the distance between each pair of neighbour images in $\text{route}_{\text{manually}}$, and the distance between each image in $\text{route}_{\text{manually}}$ and a randomly selected image in the training set. Fig 6.5 Right is the distribution of the difference of two distances, i.e. the distance between each pair of neighbour points in $\text{path}_{\text{manually}}$, and the distance between each point in $\text{path}_{\text{manually}}$ and a latent space representation of randomly selected image in the training set. This two graphs indicate that both in the real frame space and latent space, the neighbour images or points are closer than random pairs, by the Euclidean metric.

Third, we evaluate the difference between $\text{route}_{\text{geodesic}}$ and $\text{route}_{\text{manually}}$, also the difference between $\text{path}_{\text{geodesic}}$ and $\text{path}_{\text{manually}}$ by their neighbour-image-distance distributions. Fig 6.6 is the distribution of distances of neighbour image pairs in the real frame space. Fig 6.7 is the distribution of distances of every neighbour image pairs in latent space. Distributions in both spaces indicate that the difference between neighbour points or images are larger for the manually selected route than $\text{route}_{\text{geodesic}}$. And the largest distance between neighbour images is almost the same for $\text{route}_{\text{geodesic}}$ and $\text{route}_{\text{manually}}$, but $\text{route}_{\text{manually}}$ contains much more neighbour image pairs that have large distance. This indicates that the Euclidean metric for measuring similarity or continuity between images may not be ideal.

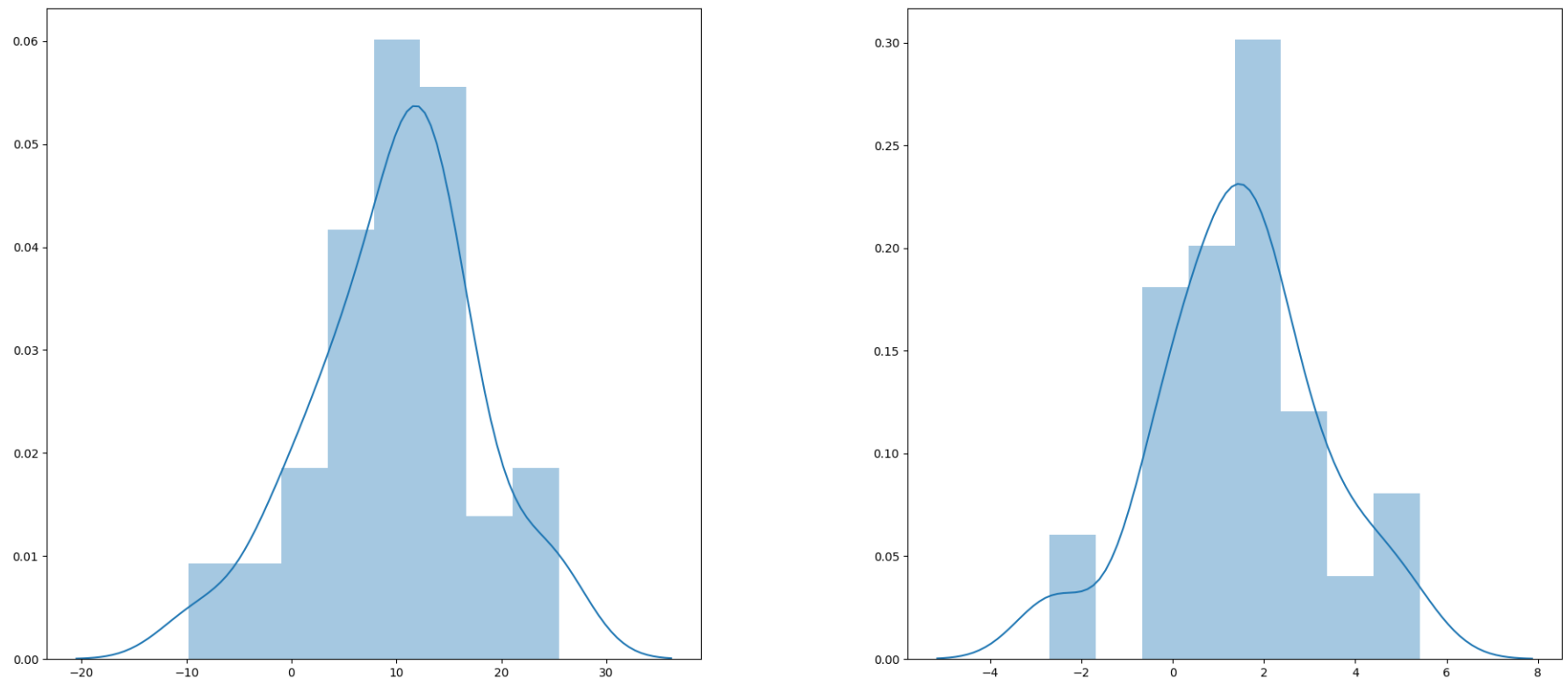

Figure 6.5: Experiment Case one: evaluation of the manually selected route. Left: the difference of two distances, i.e. the distance between each pair of neighbour images in $\text{route}_{\text{manually}}$, and the distance between each image in $\text{route}_{\text{manually}}$ and a randomly selected image in the training set; Right: the difference of two distances, i.e. the distance between each pair of neighbour points in $\text{path}_{\text{manually}}$, and the distance between each point in $\text{path}_{\text{manually}}$ and a latent space representation of randomly selected image in the training set.

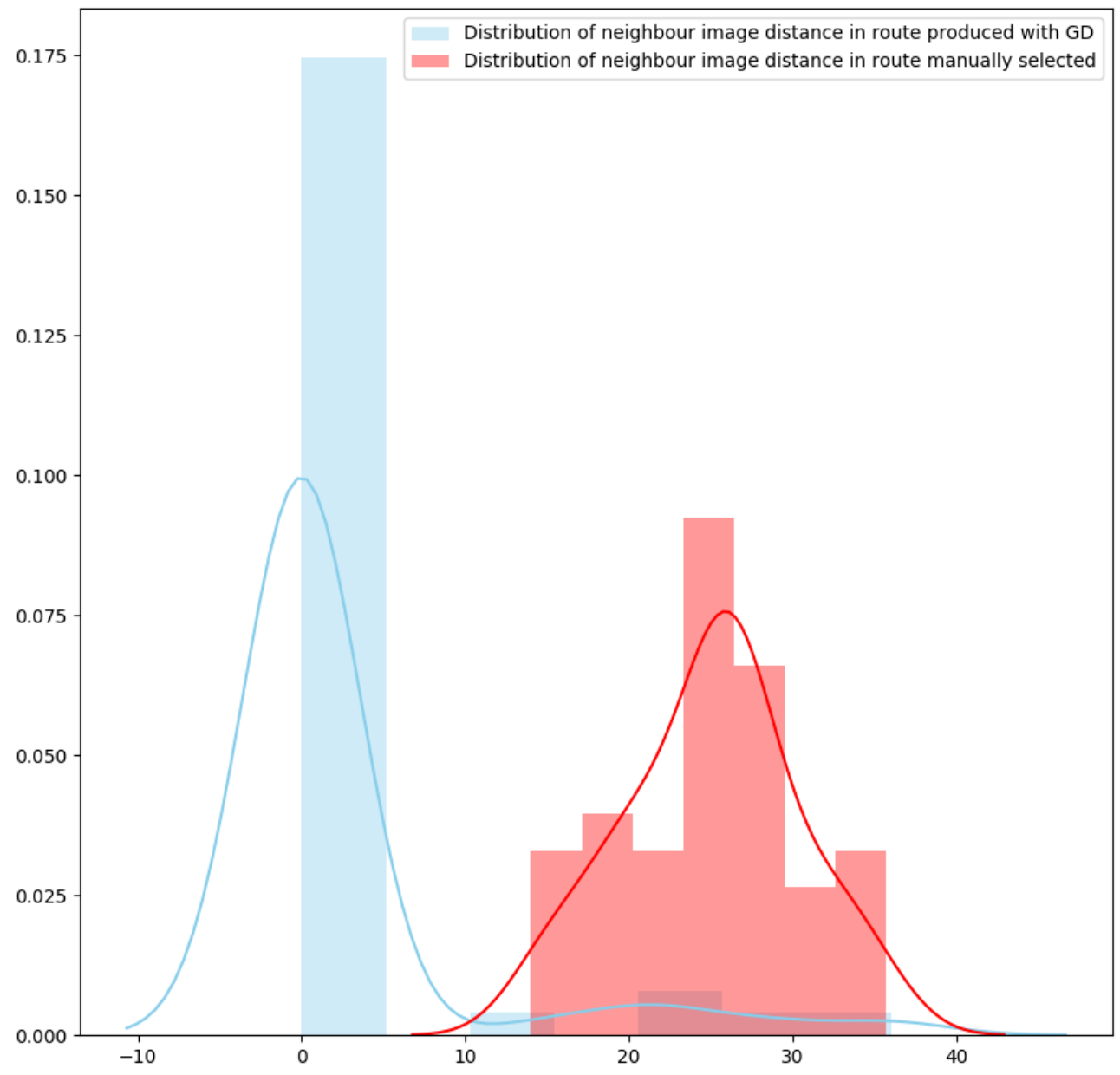

Figure 6.6: Experiment Case one: the distribution of distances of neighbour image pairs in real frame space. Blue: the distribution of distances of neighbour image pairs in $\text{route}_{\text{geodesic}}$ ; Red: the distribution of distances of neighbour image pairs in $\text{route}_{\text{manually}}$.

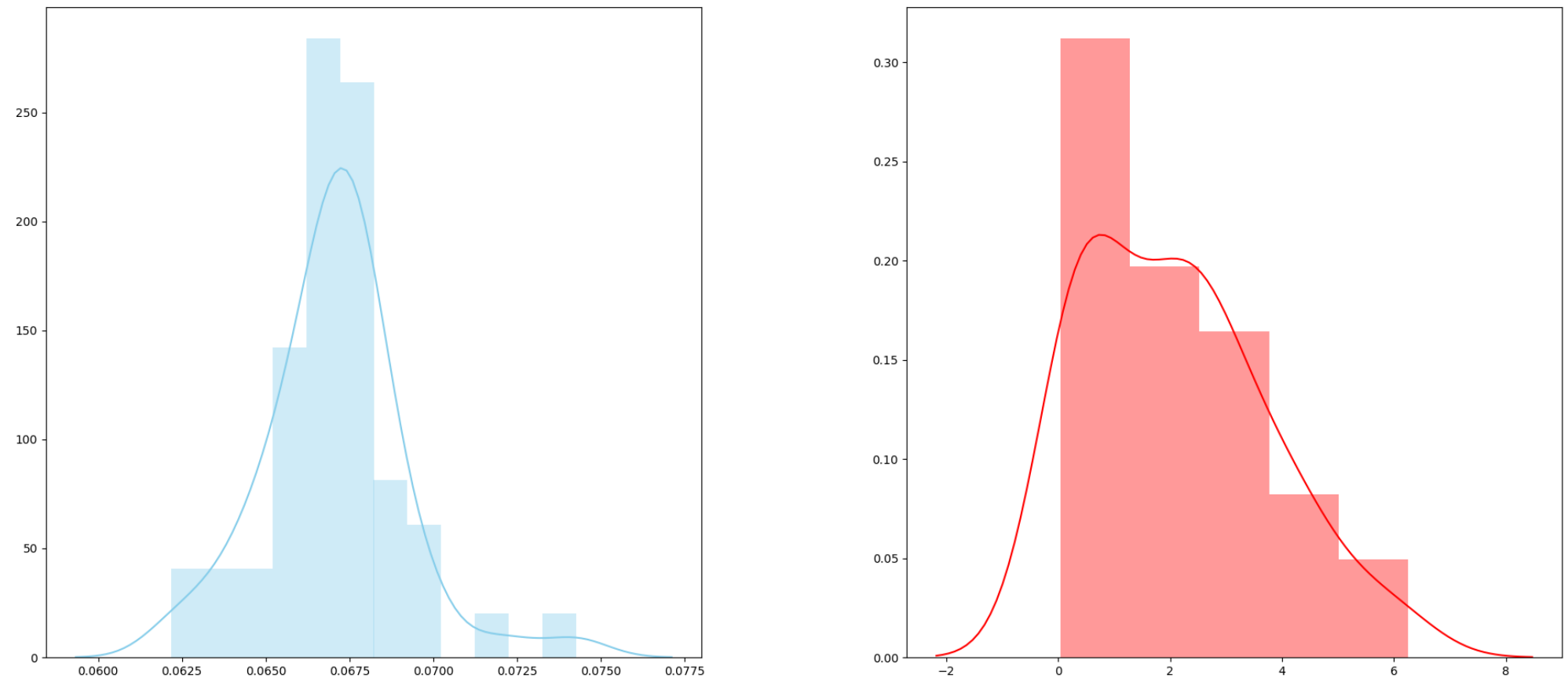

Figure 6.7: Experiment Case one: the distribution of distances of neighbour point pairs in the latent space. Left: the distribution of distances of neighbour point pairs in $\text{path}_{\text{geodesic}}$; Right: the distribution of distances of neighbour point pairs in $\text{path}_{\text{manually}}$.

## 6.2 Case two

**Starting scene: the 300-th image, the kitchen; Destination: the 650-th image, the bedroom.**

### 6.2.1 Route producing

Following the same procedure as Case one, we initialize a curve and update it with Gradient Descent method. Fig 6.8 top is the route with generated images. Fig 6.8 bottom is a route of real frames which are the most similar ones with the corresponding generated images, as is explained in 5.3.3 and 1.3.4.

The route consists of three scenes, through the kitchen to meeting room and then to the bedroom. The route is reasonable but not short and continuous.

We try to figure out if there exists such a path in the latent space that have a corresponding continuous route in reality. Fig 6.9 is the ideal route selected manually, and Fig 6.10 is the reconstructed route that produced by feeding the images of the manually selections into the trained encoder and decoder. This case confirms the same thing as Case one, that a useful route exists, because it can be selected manually, and represented by the VAE.

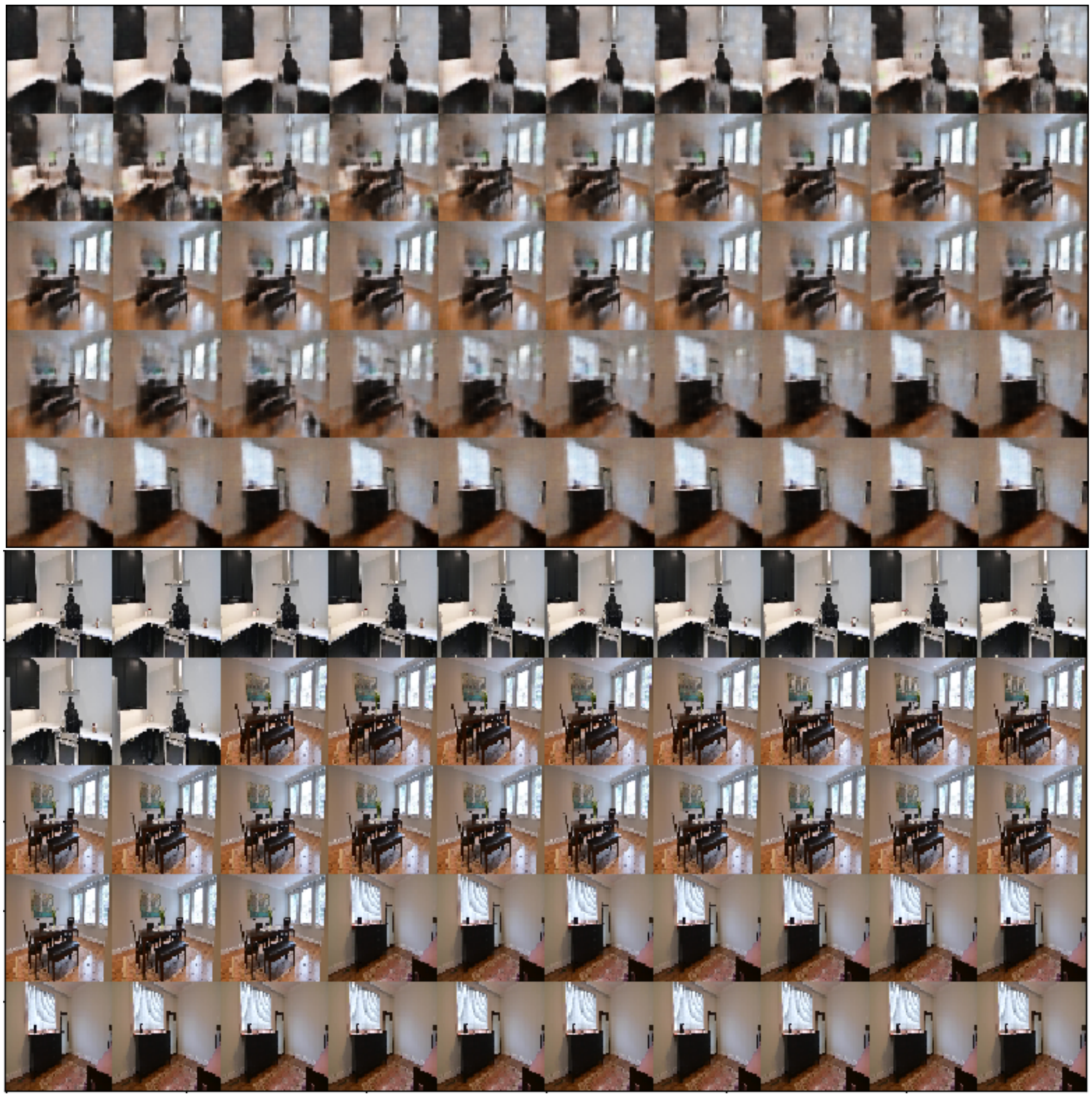

Figure 6.8: Experiment Case two: visualization of the route. The procedure for selecting such a path is explained in detail in Chapter 5. Top: route of generated images; Bottom: route of real frames, each image is the most similar real frames with the generated one, by the squared Euclidean distance similarity metric.

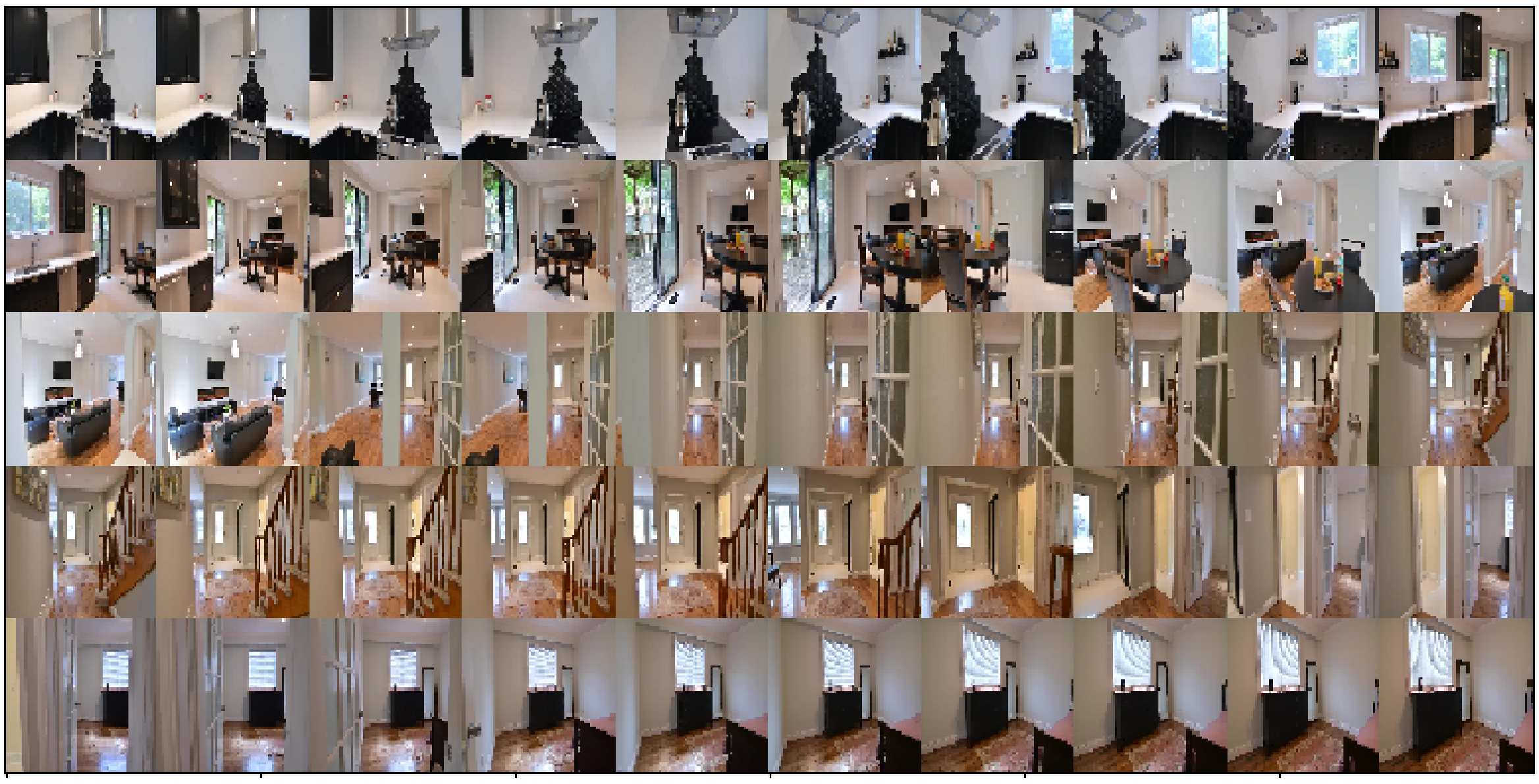

Figure 6.9: Experiment Case two: the ideal route selected manually. In this route, each neighbour image pair is continuous geographically.

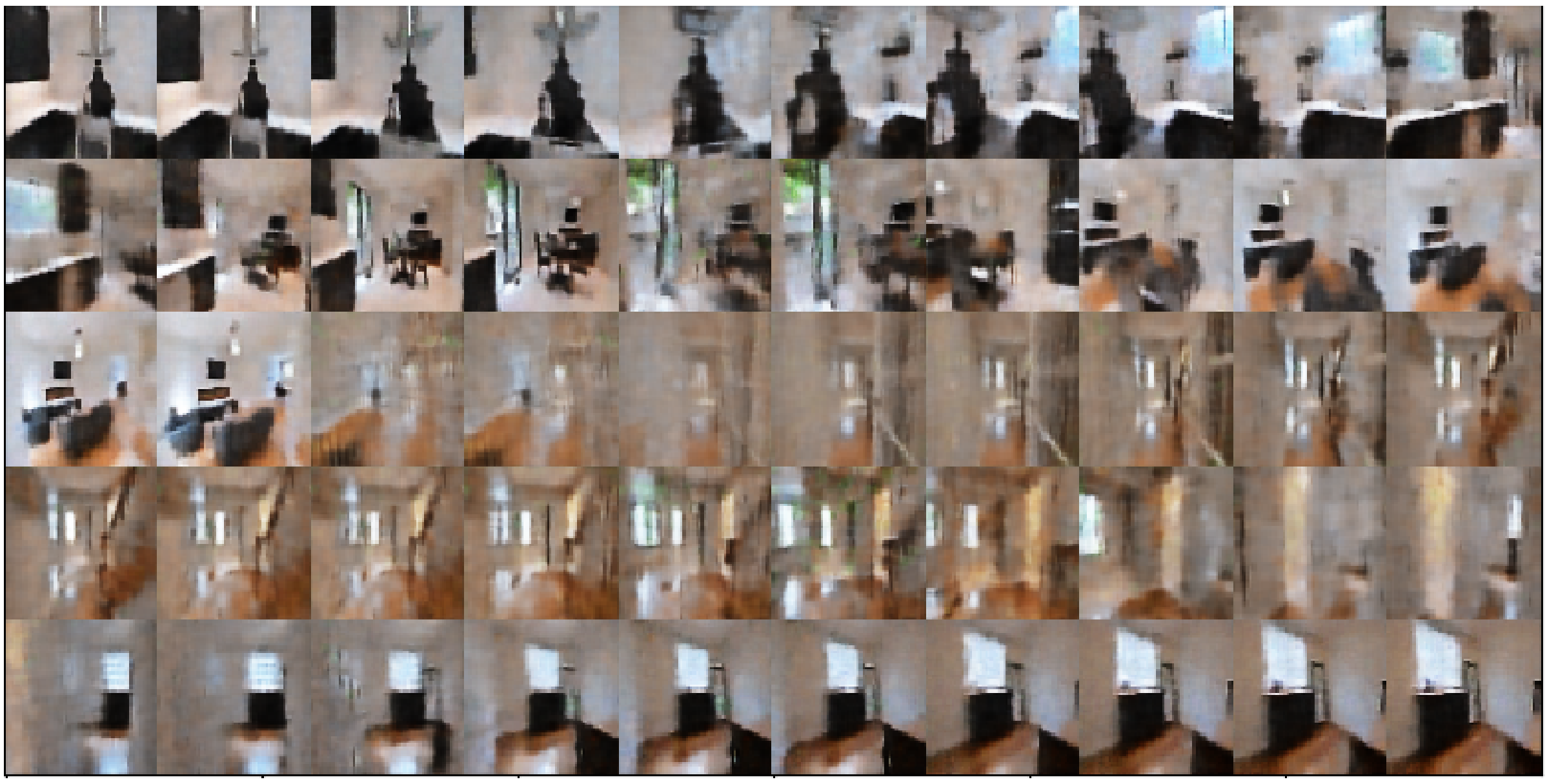

Figure 6.10: Experiment Case two: reconstructed route from an ideal route selected manually. It is produced by feeding each image along the route in Fig 6.9 through encoder and decoder.

### 6.2.2 Evaluation

First, the number of image categories in $\text{route}_{\text{geodesic}}$ over that in $\text{route}_{\text{manually}}$ is 3/50, as indicated by Fig 6.8 Bottom and Fig 6.9.

Second, we evaluate the manually selected route, and they give the same indication as Case one. Fig 6.11 indicates that both in the real frame space and latent space, the neighbour images or points are closer than random pairs, by the Euclidean metric.

Third, we evaluate the difference between $\text{route}_{\text{geodesic}}$ and $\text{route}_{\text{manually}}$, also the difference between $\text{path}_{\text{geodesic}}$ and $\text{path}_{\text{manually}}$ by their neighbour-image-distance distributions. Also the same indication as Case one is shown from Fig 6.12 and Fig 6.13..

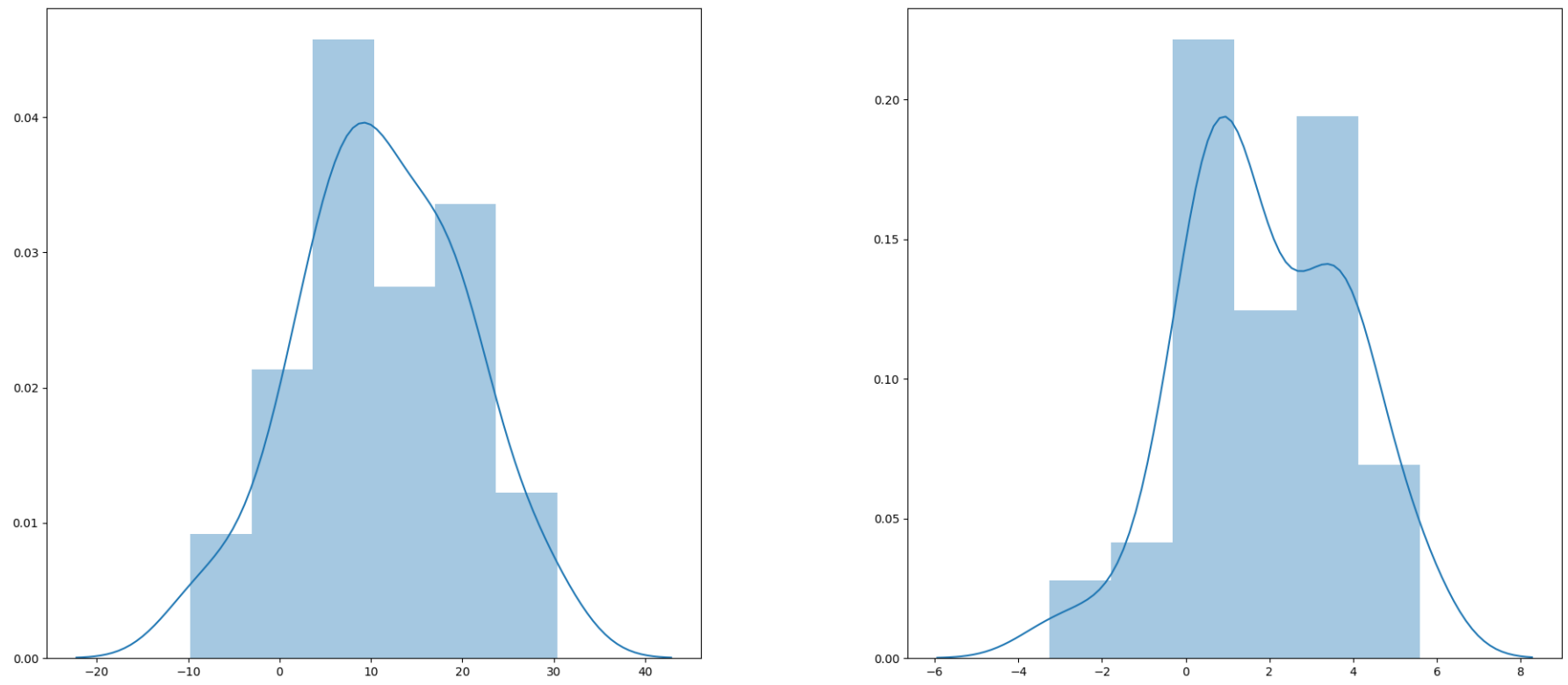

Figure 6.11: Experiment Case two: evaluation of the manually selected route. Left: the difference of two distances, i.e. the distance between each pair of neighbour images in $\text{route}_{\text{manually}}$, and the distance between each image in $\text{route}_{\text{manually}}$ and a randomly selected image in the training set; Right: the difference of two distances, i.e. the distance between each pair of neighbour points in $\text{path}_{\text{manually}}$, and the distance between each point in $\text{path}_{\text{manually}}$ and a latent space representation of randomly selected image in the training set.

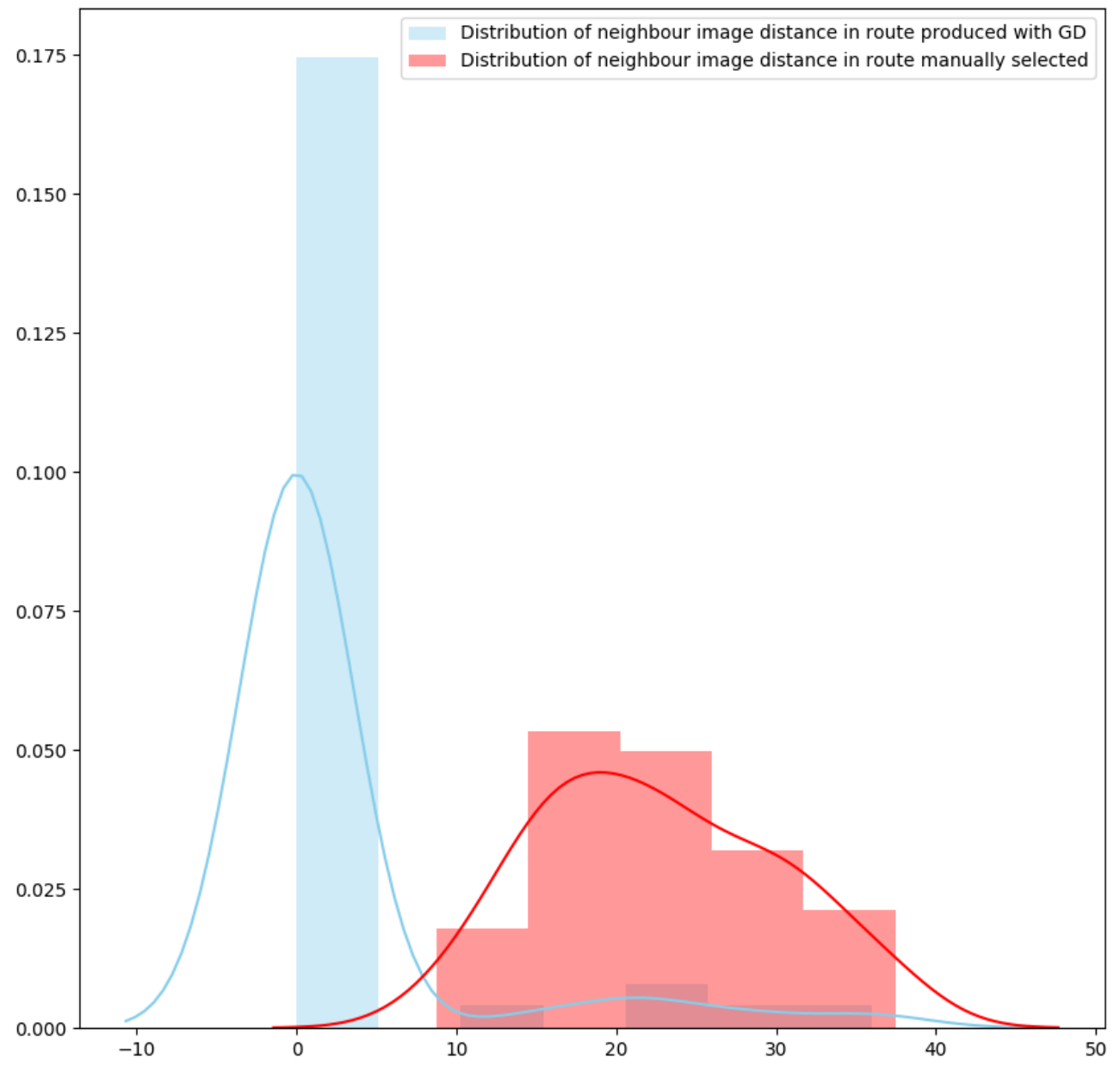

Figure 6.12: Experiment Case two: the distribution of distances of neighbour image pairs in real frame space. Blue: the distribution of distances of neighbour image pairs in $\text{route}_{\text{geodesic}}$ ; Red: the distribution of distances of neighbour image pairs in $\text{route}_{\text{manually}}$.

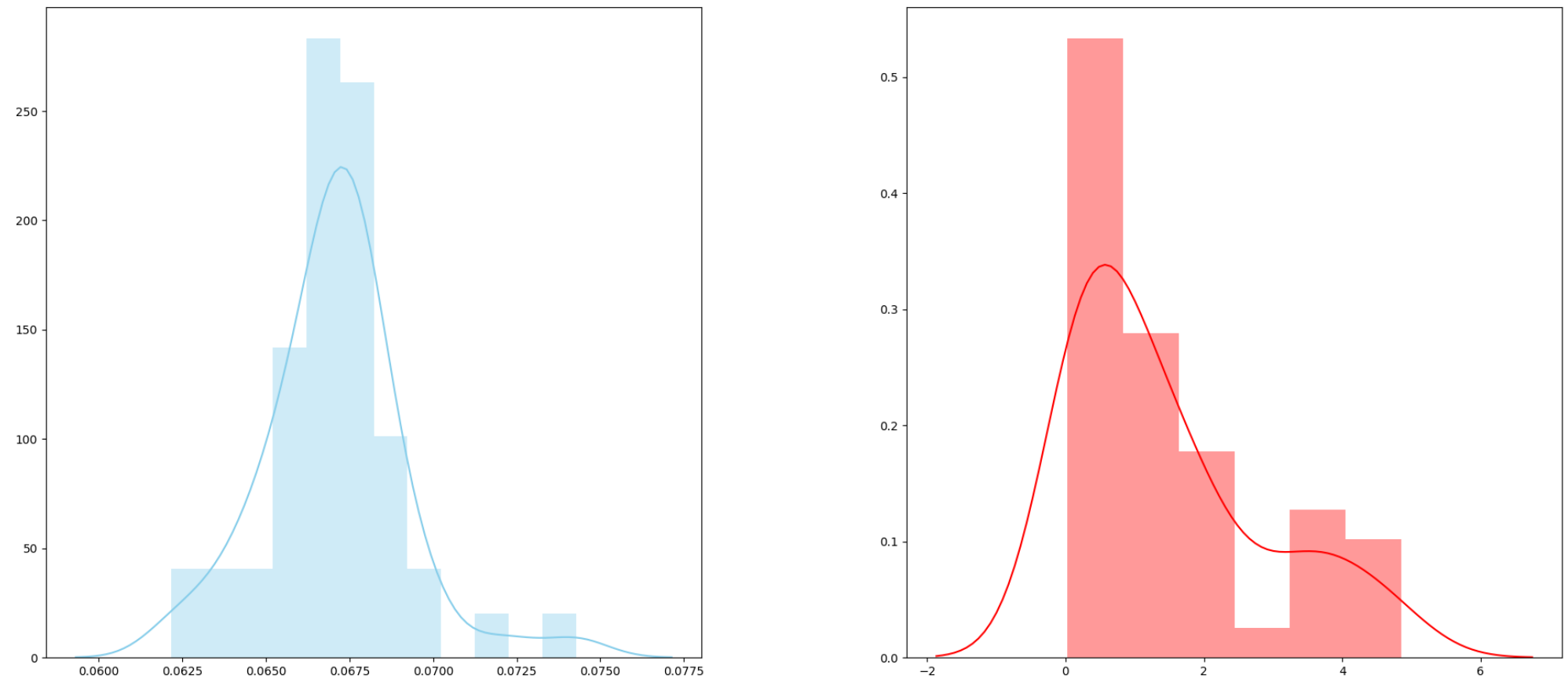

Figure 6.13: Experiment Case two: the distribution of distances of neighbour point pairs in the latent space. Left: the distribution of distances of neighbour point pairs in $\text{path}_{\text{geodesic}}$; Right: the distribution of distances of neighbour point pairs in $\text{path}_{\text{manually}}$.

## 6.3 Case three

**Starting scene: the 300-th image, the kitchen; Destination: the 750-th image, the bathroom.**

### 6.3.1 Route producing

Now we show a more extreme case in which the starting room is quite a distance from the destination, but the appearance of the two are quite similar compared with other rooms.

All the procedures are the same as last two cases, so we will not repeat them here. But the outcome is worse for this case. In Fig 6.14, the route only consists of the starting scene and the destination scene. In Fig 6.15 and 6.15 we show that the ideal route that is able to lead the robot from kitchen to the bathroom exists, because it can be selected manually, and represented by the VAE.

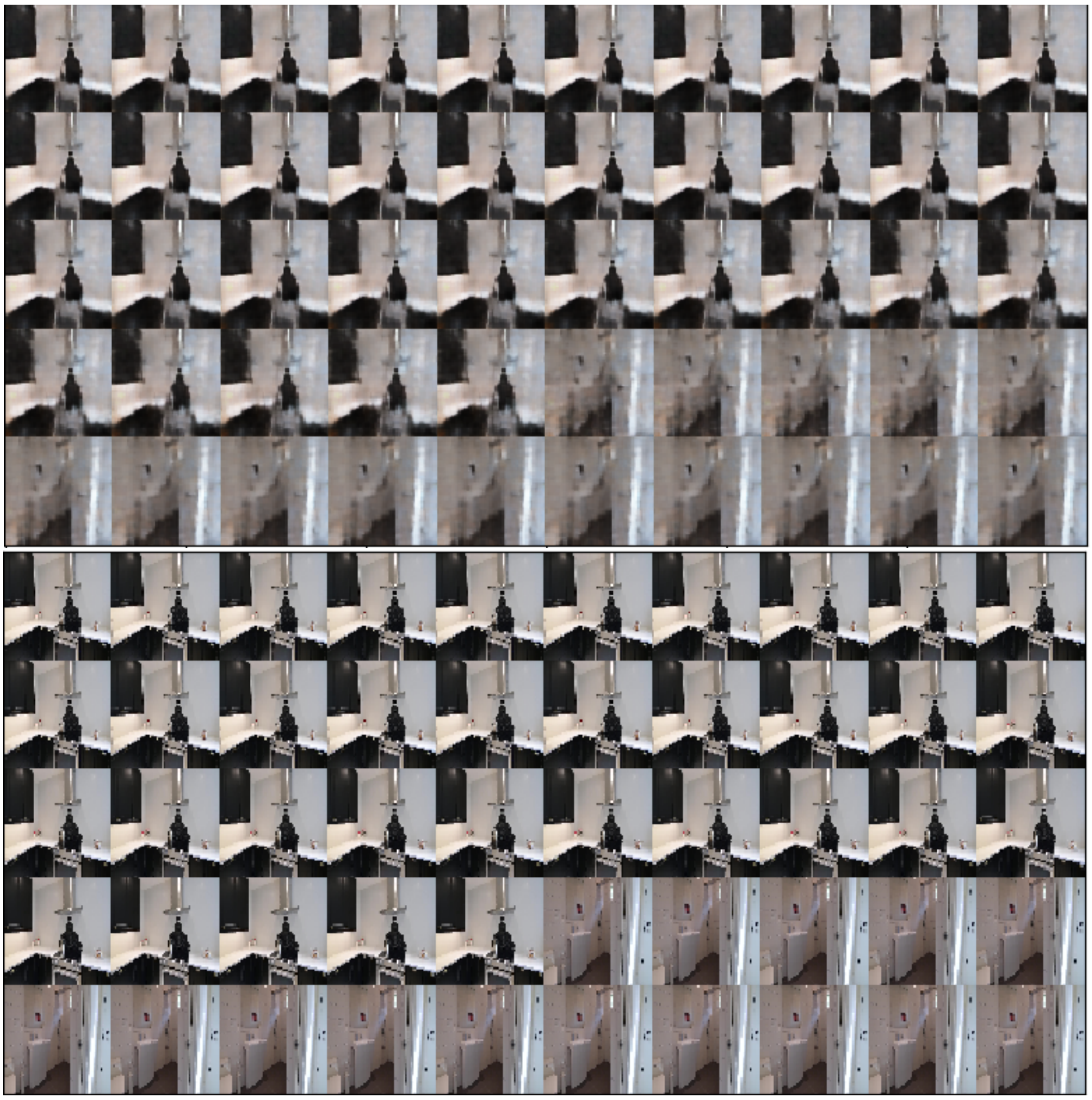

Figure 6.14: Experiment Case three: visualization of the route. The procedure for selecting such a path is explained in detail in Chapter 5. Top: route of generated images; Bottom: route of real frames, each image is the most similar real frames with the generated one, by the squared Euclidean distance similarity metric.

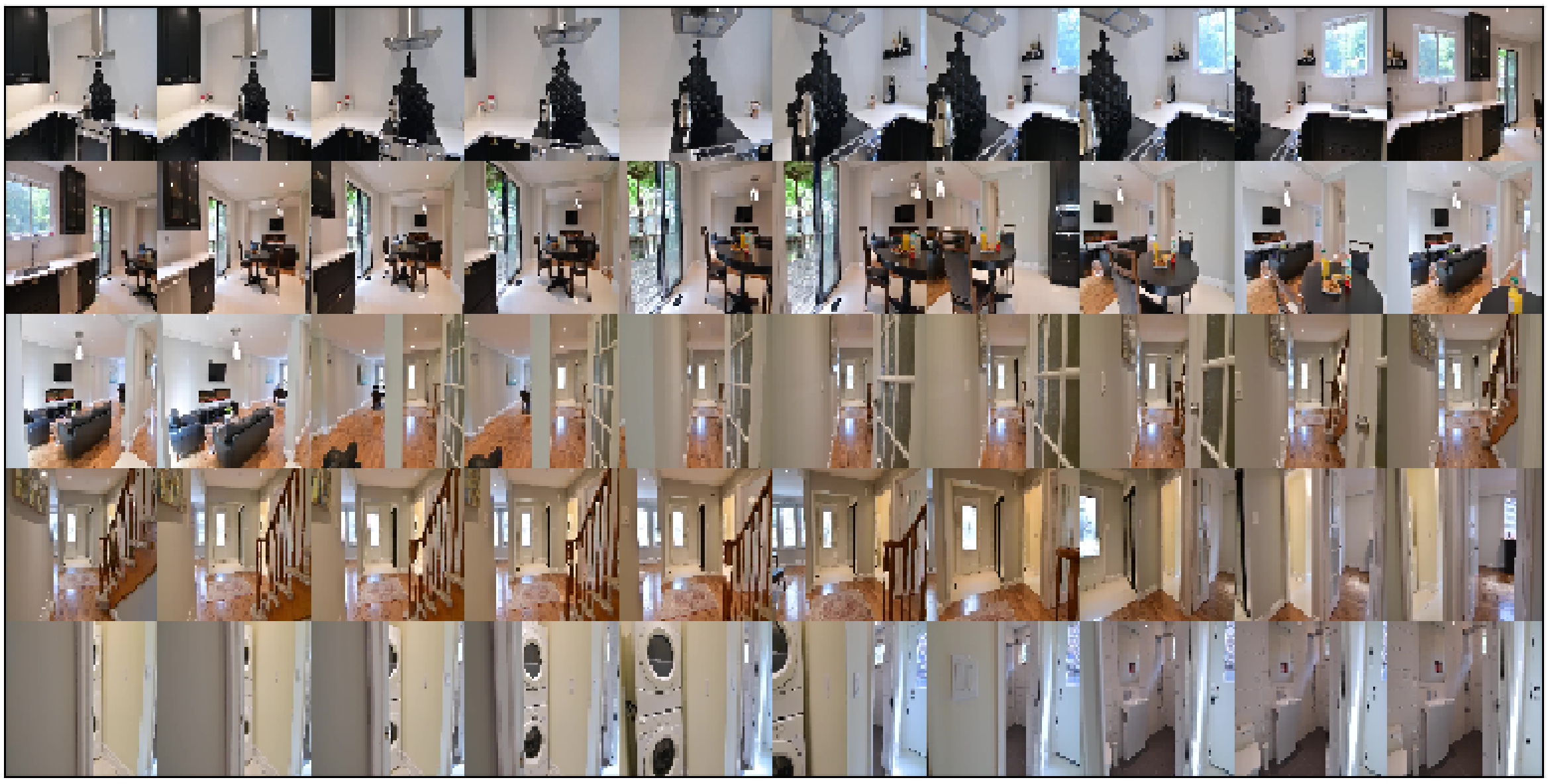

Figure 6.15: Experiment Case three: the ideal route selected manually. In this route, each neighbour image pair is continuous geographically.

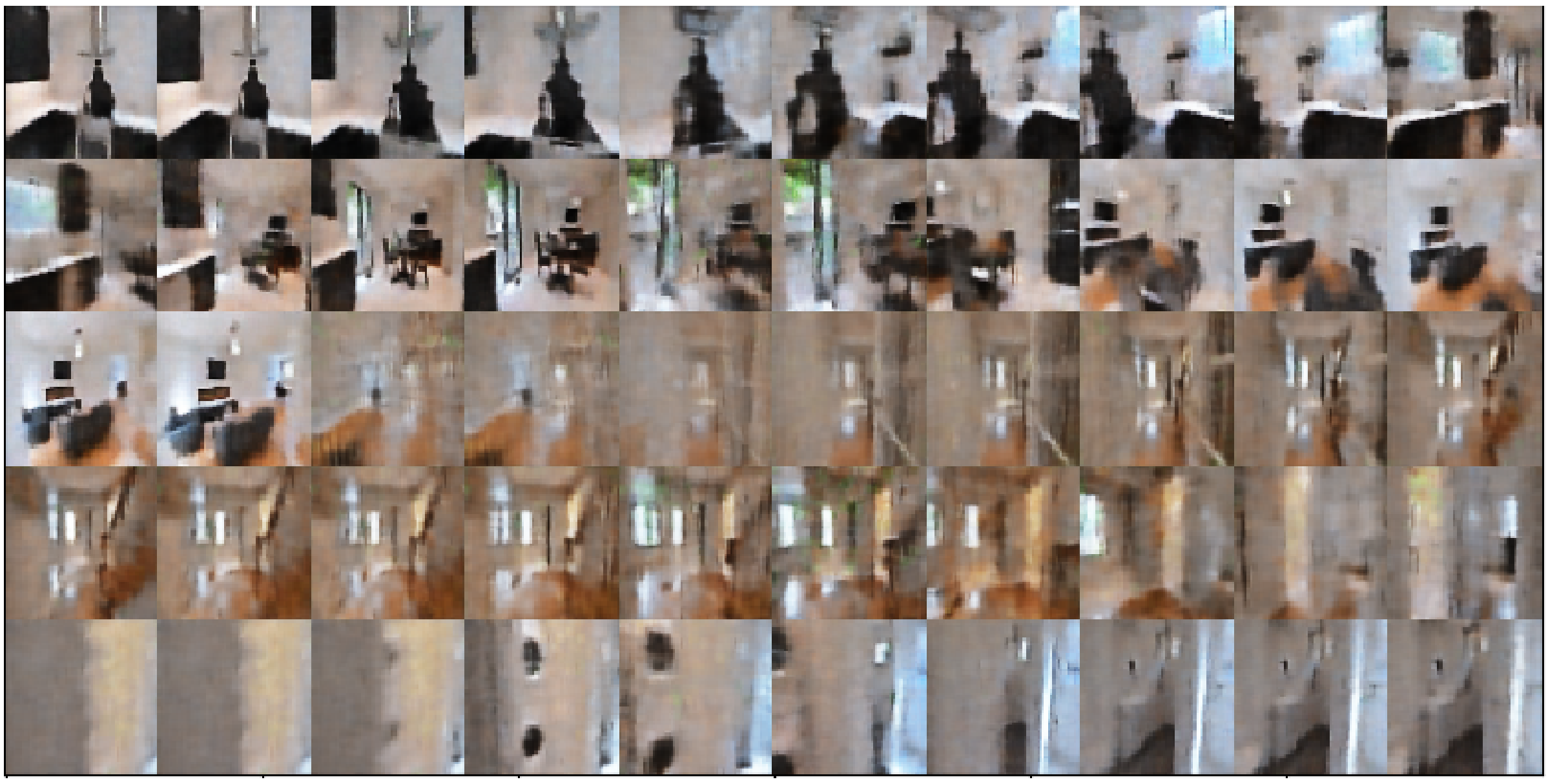

Figure 6.16: Experiment Case three: reconstructed route from an ideal route selected manually. It is produced by feeding each image along the route in Fig 6.15 through encoder and decoder.

### 6.3.2 Evaluation

First, the number of image categories in $\text{route}_{\text{geodesic}}$ over that in $\text{route}_{\text{manually}}$ is only 2/50, as indicated by Fig 6.14 Bottom and Fig 6.15, which is very bad.

Second, we evaluate the manually selected route. Fig 6.17 indicates different things from the previous two cases: while the difference of the distance between each pair of neighbour images in $\text{route}_{\text{manually}}$, and the distance between each image in $\text{route}_{\text{manually}}$ and a randomly selected image in the training set is above zero, the difference of two distances, i.e. the distance between each pair of neighbour points in $\text{path}_{\text{manually}}$, and the distance between each point in $\text{path}_{\text{manually}}$ and a latent space representation of randomly selected image in the training set is around near zero. This means that the learned map is useless for providing path between locations with very similar appearance.

Third, we evaluate the difference between $\text{route}_{\text{geodesic}}$ and $\text{route}_{\text{manually}}$, also the difference between $\text{path}_{\text{geodesic}}$ and $\text{path}_{\text{manually}}$ by their neighbour-image-distance distributions. Also the same indication as Case one and two is shown from the Fig 6.18 and Fig 6.19.

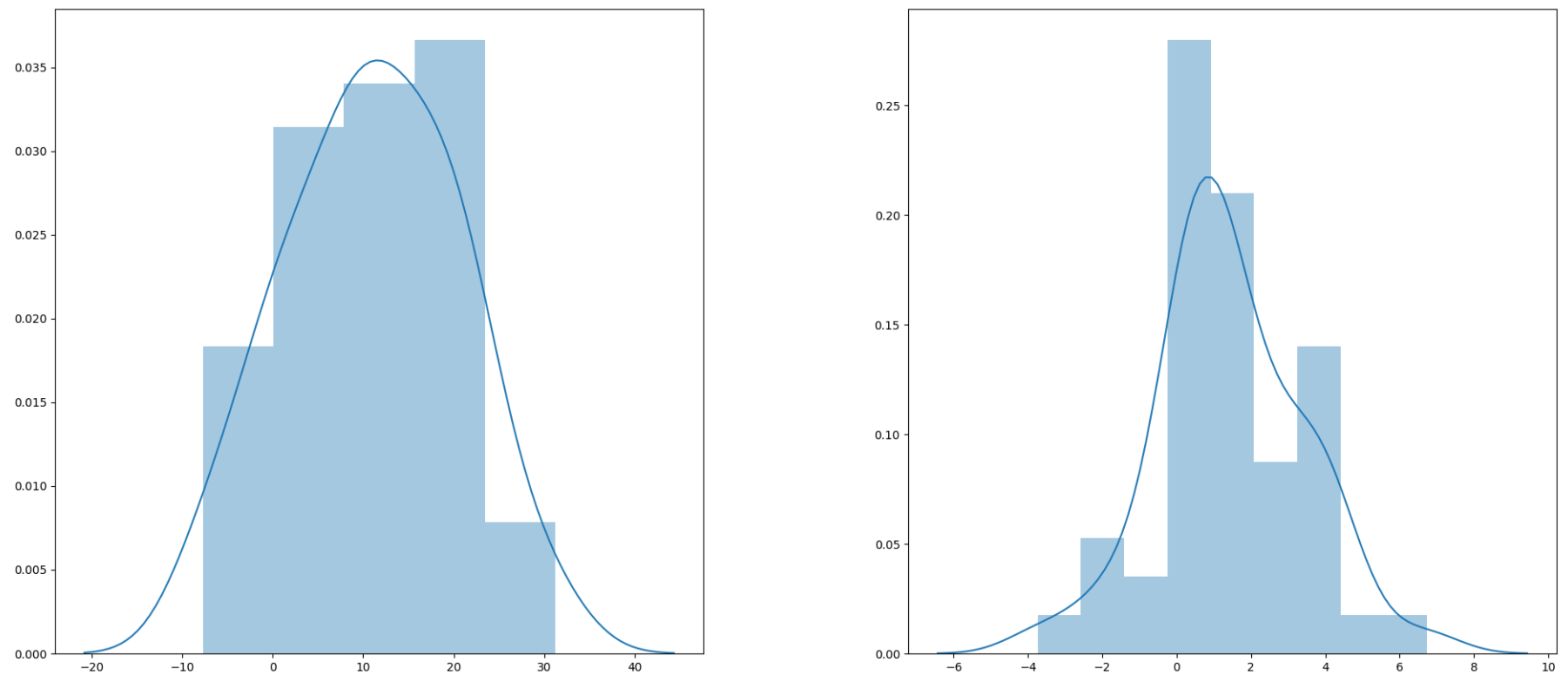

Figure 6.17: Experiment Case three: evaluation of the manually selected route. Left: the difference of two distances, i.e. the distance between each pair of neighbour images in $\text{route}_{\text{manually}}$, and the distance between each image in $\text{route}_{\text{manually}}$ and a randomly selected image in the training set; Right: the difference of two distances, i.e. the distance between each pair of neighbour points in $\text{path}_{\text{manually}}$, and the distance between each point in $\text{path}_{\text{manually}}$ and a latent space representation of randomly selected image in the training set.

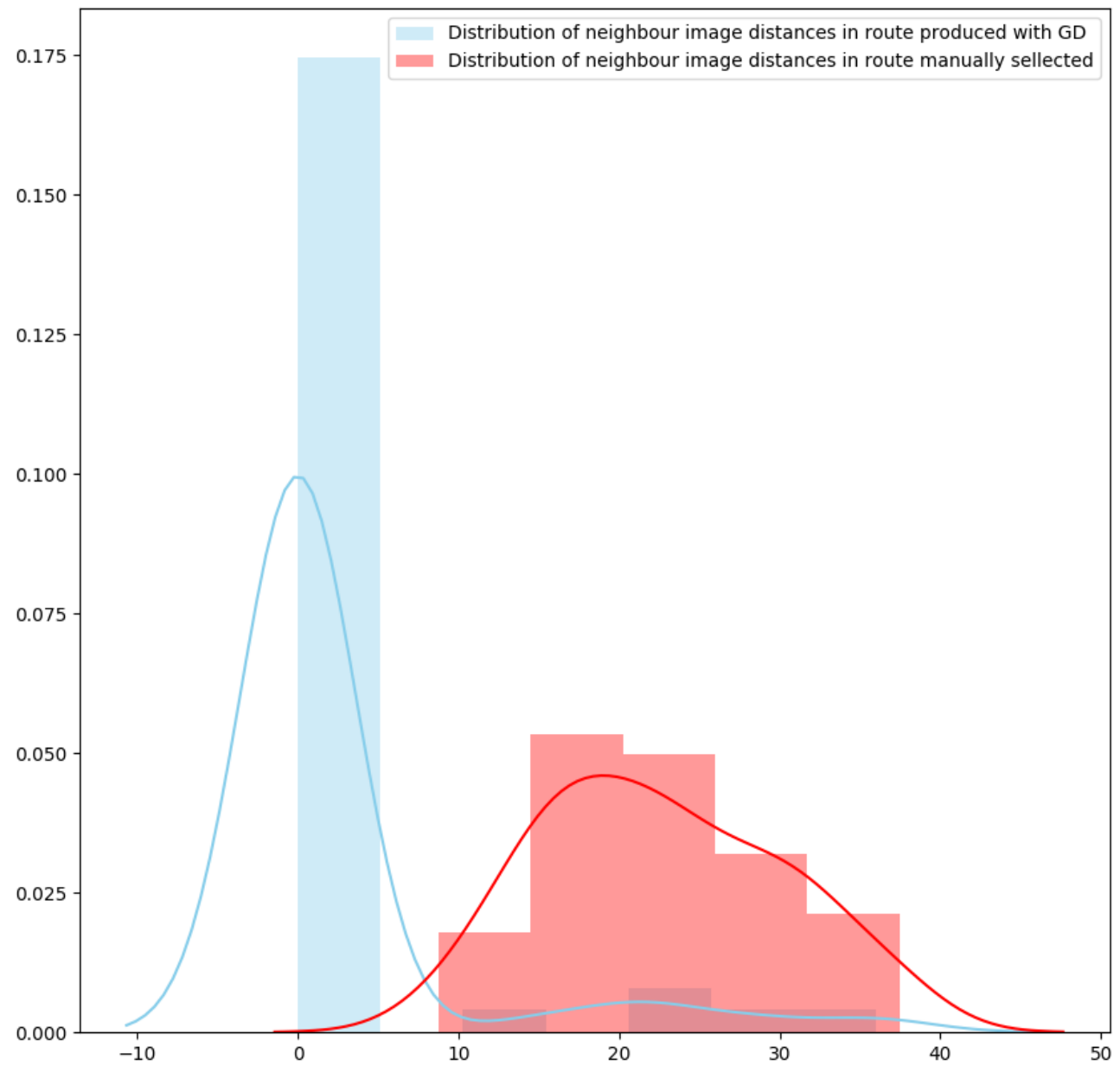

Figure 6.18: Experiment Case three: the distribution of distances of neighbour image pairs in real frame space. Blue: the distribution of distances of neighbour image pairs in $\text{route}_{\text{geodesic}}$ ; Red: the distribution of distances of neighbour image pairs in $\text{route}_{\text{manually}}$.

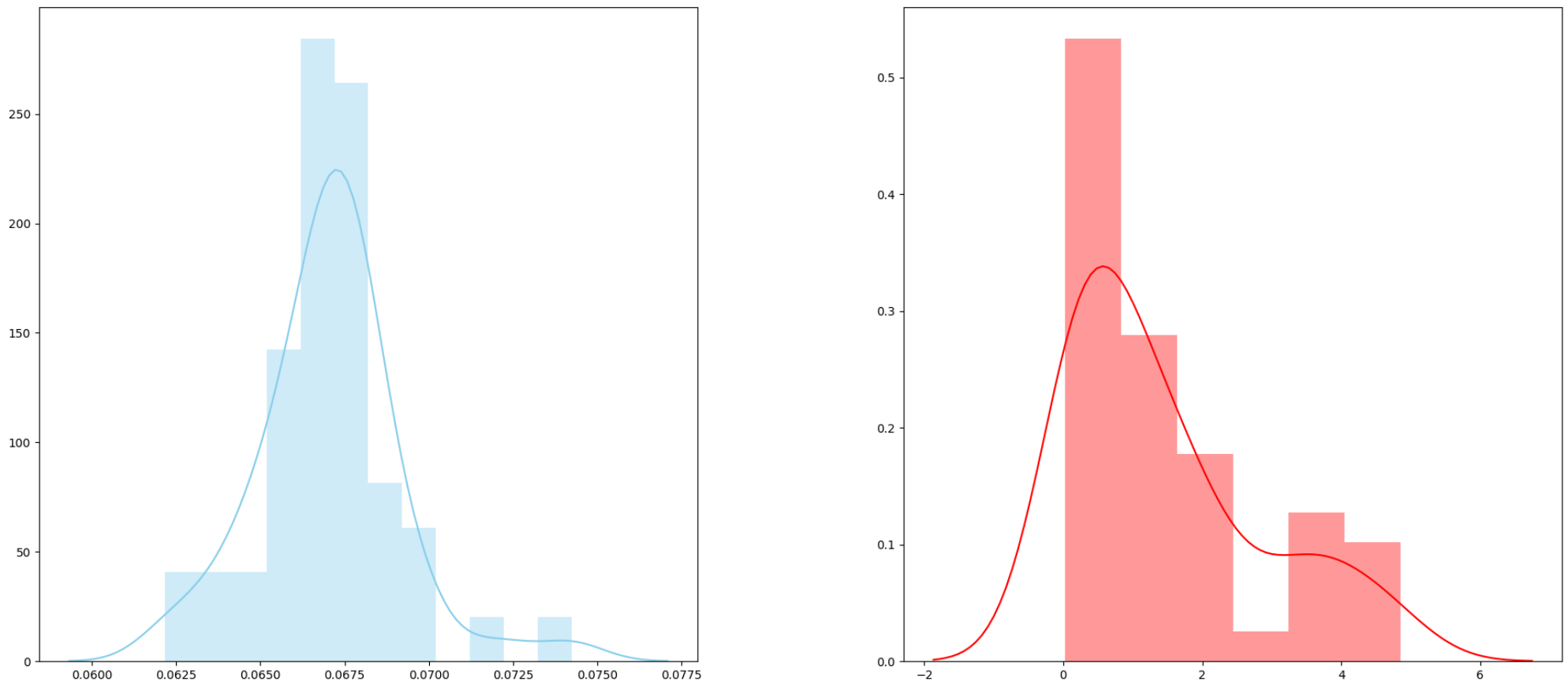

Figure 6.19: Experiment Case three: the distribution of distances of neighbour point pairs in the latent space. Left: the distribution of distance of neighbour point pairs in $\text{path}_{\text{geodesic}}$; Right: the distribution of distances of neighbour point pairs in $\text{path}_{\text{manually}}$.

# Chapter 7

# Conclusion and future work

## 7.1 Conclusion

The aim of this work is to use VAE to learn a representation map for the robot navigation. We make a hypothesis in Chapter 5.2 that the learned representation map is continuous and complete, such that we choose geodesic as the path. We implement this algorithm, but find in our experimental results that the resulting route is not satisfactory. The routes produced in the experiments are not ideal, for they consist of several discontinuous image frames along the ideal routes, so that the route could not be followed by a robot with computer vision techniques in practice. In our evaluation, we propose two reasons related to the hypothesis for our failure to automatically find continuous routes:

1. One reason may be related to the VAE method. It seems that the representation learned by the VAE captures global structures but discards details. It can be inferred from the visualization of learned manifold slices in Fig 4.2, that many reconstructed images are blurry, and are combinations of two scenes that are from close rooms, in which we can't figure out the exact locations they represent. So the produced routes lack continuity for some locations.

2. Another point for constructing the ideal route may be related to the similarity metric used in the optimization function of path planning algorithm. Fig 6.6, Fig 6.12, Fig 6.18 and especially Fig 6.17 indicate that the similarity metric is sub-optimal, as the non-ideal route has smaller difference between neighbour images. This is because a shift of an image doesn't change much geographically, but it change a lot in the Euclidean loss function. Especially in experiment case three in Chapter 6.3, the starting and ending images are distant from each other geographically, but their representations are relatively close in the learned manifold, because of the similar layout of the room, as a result the path selecting algorithm is not working at all.

## 7.2 Future work

Future work to improve the route can be in two directions:

1. We could try other generative models which are better at reconstructing the details to learn the representation map. For example, VAE-GANs models, as discussed in Related work overview chapter. Many [Larsen et al., 2015] [Dosovitskiy and Brox, 2016] [Mescheder et al., 2017] [Rosca et al., 2017] reported improvement by different evaluation metrics.

2. We could adjust the similarity metric. For example,Dosovitskiy and Brox [2016] added loss in feature space and adversarial loss to the Euclidean loss in the image space, and developed a class of loss function called deep perceptual similarity metrics. So further work of this project can be devoted to developing a more suitable optimization function with consideration of feature space for the path selecting algorithm. Also the "aliasing" problem associated with similar appearance of distinct locations should be taken into consideration.

# Appendices

# Appendix A

# Structure of the model network

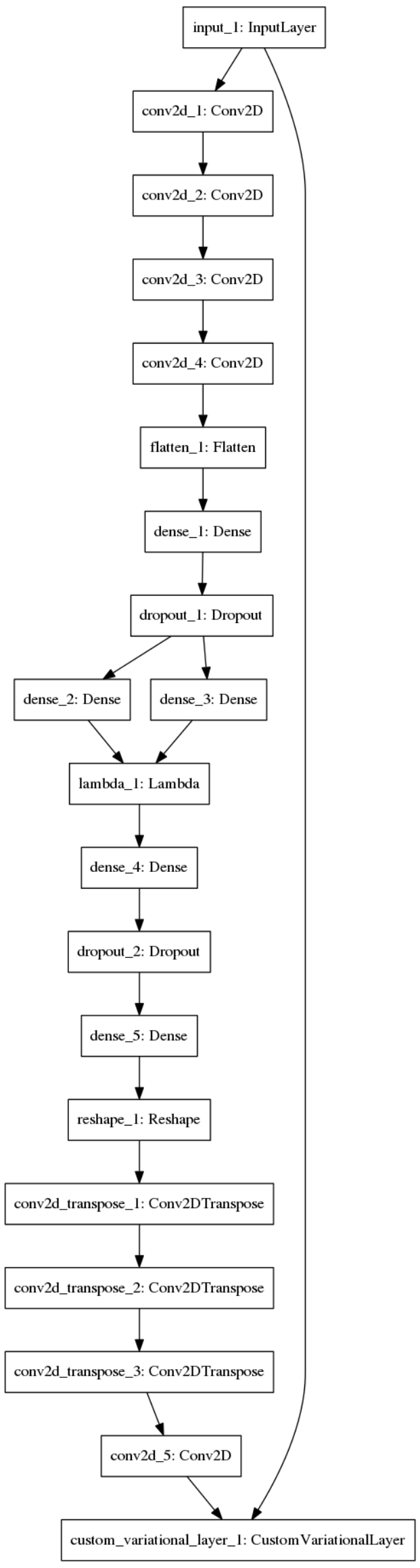

Figure A.1: model structure

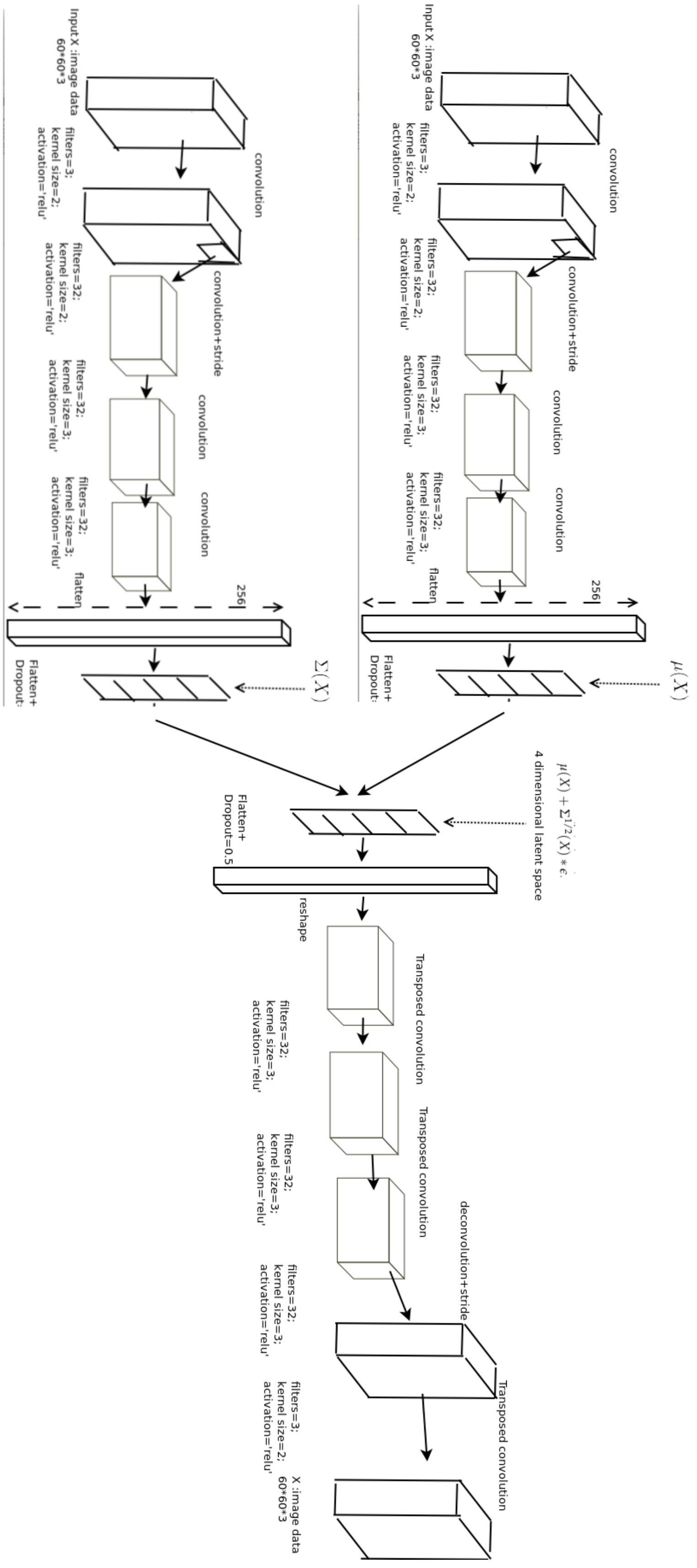

Figure A.2: model structure, and network details

# Appendix B

# visualization of Variational Autoencoder with 2 D latent space

Note, that this project doesn't adopt the 2 dimension latent space model, and this appendix provide the experimental evidence why we didn't it.

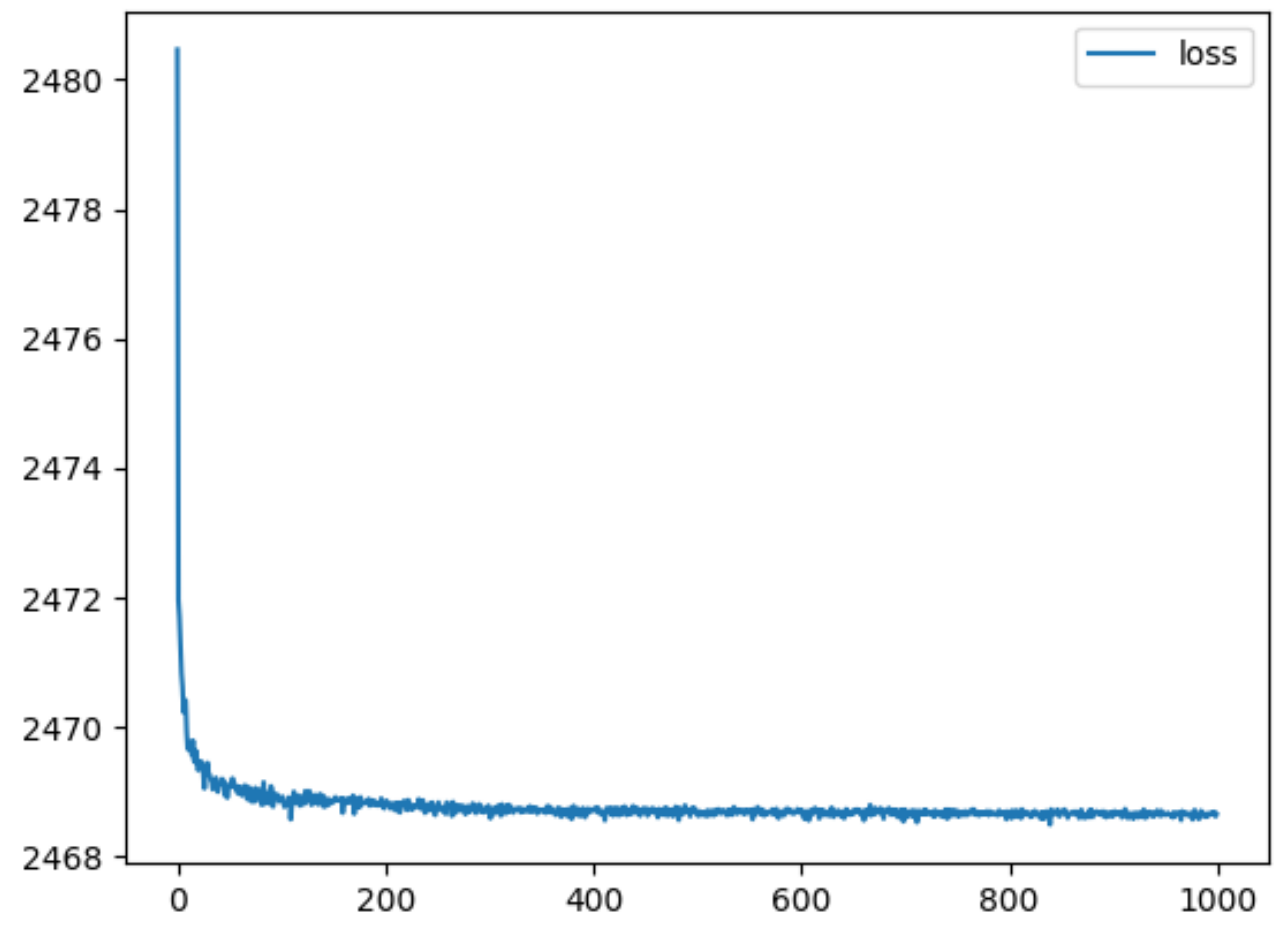

Figure B.1: The loss value for a 2d latent space model through the training. The loss function in a Variational Autoencoder is the negative lower bound of the log probability of the training data.

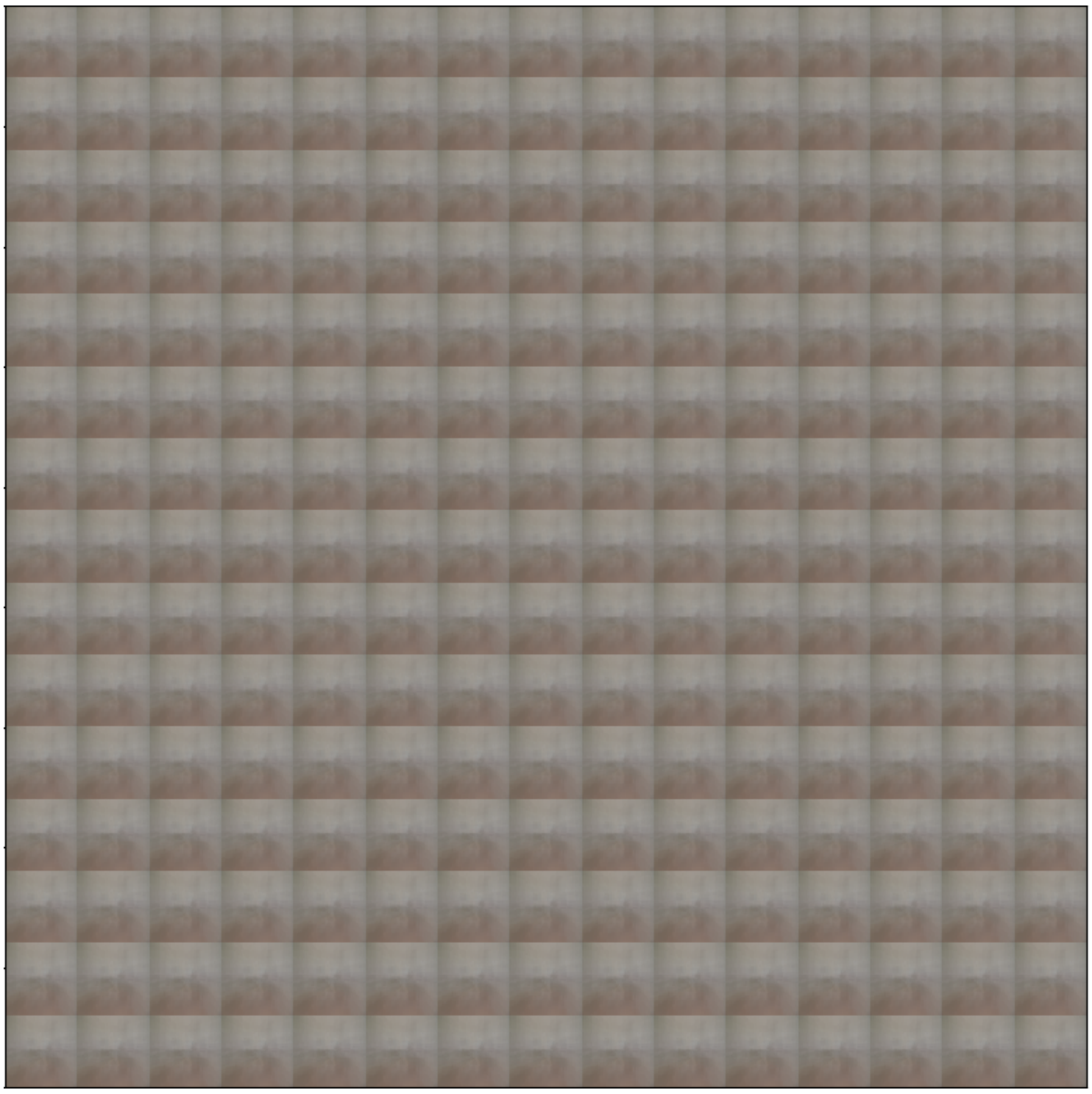

Figure B.2: Visualization of 2 D manifold with a Variatioinal Autoencoder Model with 2 D latent space.

# Appendix C

# Visualize slices of 4 D manifold

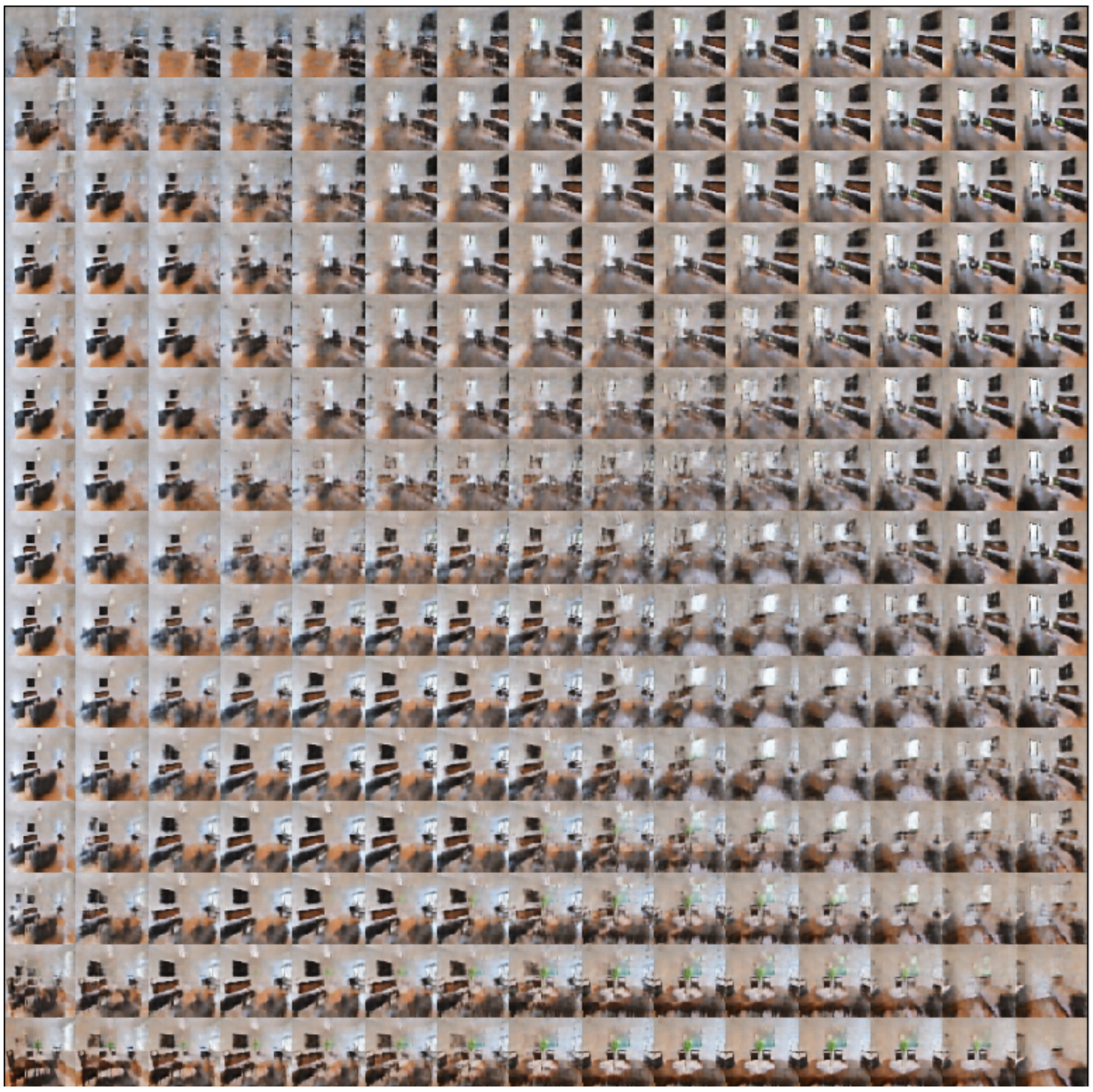

Figure C.1: A slice of the latent space manifold.
Sampling latent variable z : We set the first two dimensions to be linearly spaced between 0.05 and 0.95, and set values for remaining dimensions to 0. So the result is a 2-D grid, in terms of the first two dimensions, linearly spaced on the square bounded by 0.05 and 0.9.

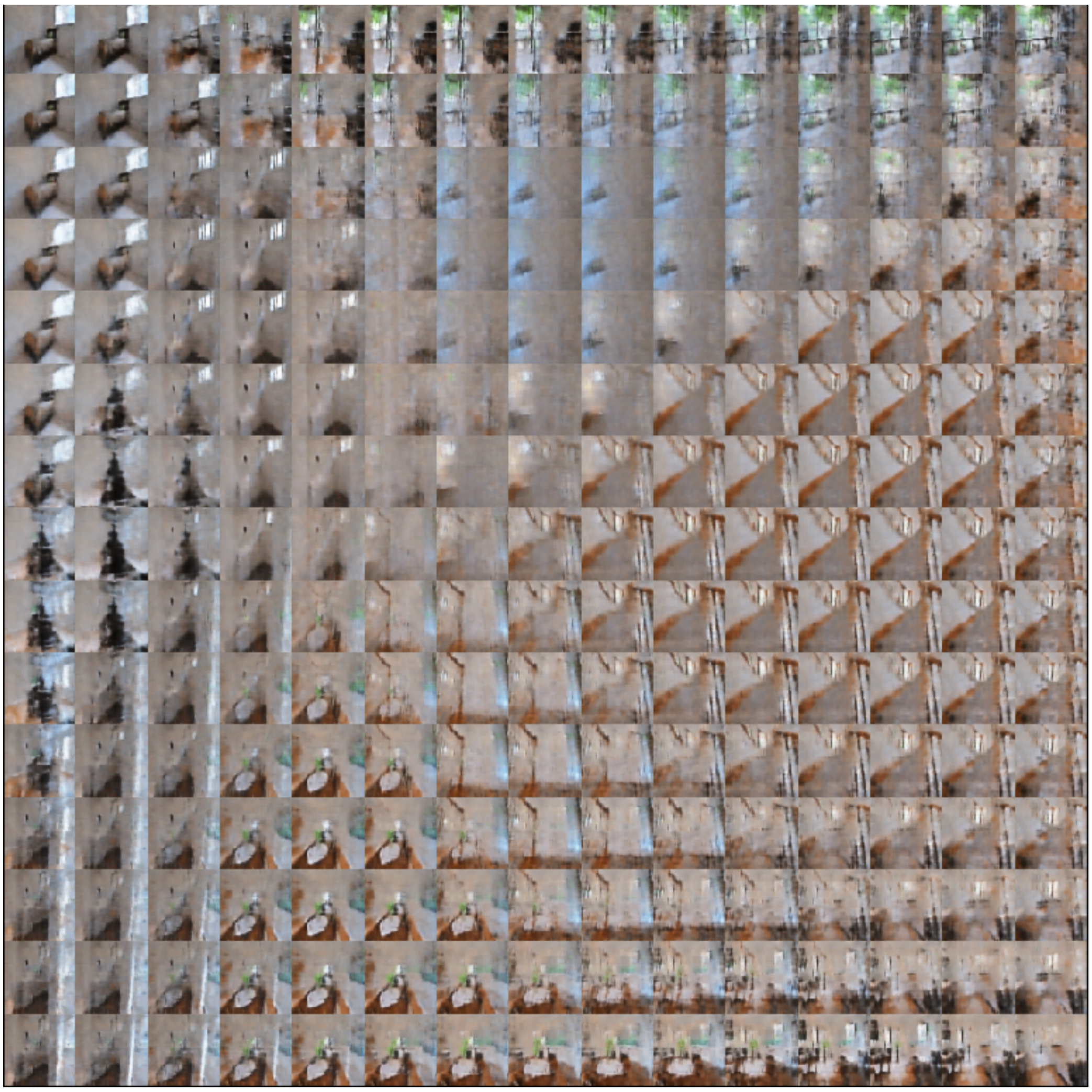

Figure C.2: A slice of the latent space manifold. Sampling latent variable z : We set the first two dimensions to be linearly spaced between 0.05 and 0.95, and set values for remaining dimensions to 0. So the result is a 2-D grid, in terms of the first two dimensions, linearly spaced on the square bounded by 0.05 and 0.95

# Bibliography

Angelo Arleo and Wulfram Gerstner. Spatial cognition and neuro-mimetic navigation: a model of hippocampal place cell activity. *Biological cybernetics*, 83(3): 287–299, 2000.

Yoshua Bengio, Aaron Courville, and Pascal Vincent. Representation learning: A review and new perspectives. *IEEE transactions on pattern analysis and machine intelligence*, 35(8):1798–1828, 2013.

Xi Chen, Diederik P Kingma, Tim Salimans, Yan Duan, Prafulla Dhariwal, John Schulman, Ilya Sutskever, and Pieter Abbeel. Variational lossy autoencoder. *arXiv preprint arXiv:1611.02731*, 2016.

Guilherme N DeSouza and Avinash C Kak. Vision for mobile robot navigation: A survey. *IEEE transactions on pattern analysis and machine intelligence*, 24(2): 237–267, 2002.

Carl Doersch. Tutorial on variational autoencoders. *arXiv preprint arXiv:1606.05908*, 2016.

Alexey Dosovitskiy and Thomas Brox. Generating images with perceptual similarity metrics based on deep networks. In *Advances in Neural Information Processing Systems*, pages 658–666, 2016.

David Filliat and Jean-Arcady Meyer. Map-based navigation in mobile robots:: I. a review of localization strategies. *Cognitive Systems Research*, 4(4):243–282, 2003.

Jorge Fuentes-Pacheco, José Ruiz-Ascencio, and Juan Manuel Rendón-Mancha. Visual simultaneous localization and mapping: a survey. *Artificial Intelligence Review*, 43(1):55–81, 2015.

Philippe Gaussier, Cédric Joulain, Stéphane Zrehen, Jean-Paul Banquet, and Arnaud Revel. Visual navigation in an open environment without map. In *Intelligent Robots and Systems, 1997. IROS'97., Proceedings of the 1997 IEEE/RSJ International Conference on*, volume 2, pages 545–550. IEEE, 1997.

Ian Goodfellow, Jean Pouget-Abadie, Mehdi Mirza, Bing Xu, David Warde-Farley, Sherjil Ozair, Aaron Courville, and Yoshua Bengio. Generative adversarial nets. In *Advances in neural information processing systems*, pages 2672–2680, 2014.

Ian Goodfellow, Yoshua Bengio, and Aaron Courville. *Deep Learning*. MIT Press, 2016. `http://www.deeplearningbook.org`.

Geoffrey E Hinton and Drew Van Camp. Keeping the neural networks simple by minimizing the description length of the weights. In *Proceedings of the sixth annual conference on Computational learning theory*, pages 5–13. ACM, 1993.

Geoffrey E Hinton, Peter Dayan, Brendan J Frey, and Radford M Neal. The" wake-sleep" algorithm for unsupervised neural networks. *Science*, 268(5214):1158–1161, 1995.

Geoffrey E Hinton, Simon Osindero, and Yee-Whye Teh. A fast learning algorithm for deep belief nets. *Neural computation*, 18(7):1527–1554, 2006.

Matthew D Hoffman, David M Blei, Chong Wang, and John Paisley. Stochastic variational inference. *The Journal of Machine Learning Research*, 14(1):1303–1347, 2013.

Antti Honkela, Tapani Raiko, Mikael Kuusela, Matti Tornio, and Juha Karhunen. Approximate riemannian conjugate gradient learning for fixed-form variational bayes. *Journal of Machine Learning Research*, 11(Nov):3235–3268, 2010.

Michael I Jordan, Zoubin Ghahramani, Tommi S Jaakkola, and Lawrence K Saul. An introduction to variational methods for graphical models. *Machine learning*, 37(2):183–233, 1999.

Cédric Joulain, Philippe Gaussier, Arnaud Revel, and B Gas. Learning to build visual categories from perception-action associations. In *Intelligent Robots and Systems, 1997. IROS'97., Proceedings of the 1997 IEEE/RSJ International Conference on*, volume 2, pages 857–864. IEEE, 1997.

Dongsung Kim and Ramakant Nevatia. Recognition and localization of generic objects for indoor navigation using functionality. *Image and Vision Computing*, 16(11):729–743, 1998.

Dongsung Kim and Ramakant Nevatia. Symbolic navigation with a generic map. *Autonomous Robots*, 6(1):69–88, 1999.

Diederik P Kingma and Max Welling. Auto-encoding variational bayes. *arXiv preprint arXiv:1312.6114*, 2013.

Diederik P Kingma, Shakir Mohamed, Danilo Jimenez Rezende, and Max Welling. Semi-supervised learning with deep generative models. In *Advances in Neural Information Processing Systems*, pages 3581–3589, 2014.

Diederik P Kingma, Tim Salimans, Rafal Jozefowicz, Xi Chen, Ilya Sutskever, and Max Welling. Improved variational inference with inverse autoregressive flow. In *Advances in Neural Information Processing Systems*, pages 4743–4751, 2016.

Benjamin Kuipers and Yung-Tai Byun. A robot exploration and mapping strategy based on a semantic hierarchy of spatial representations. *Robotics and autonomous systems*, 8(1-2):47–63, 1991.

Tejas D Kulkarni, William F Whitney, Pushmeet Kohli, and Josh Tenenbaum. Deep convolutional inverse graphics network. In *Advances in Neural Information Processing Systems*, pages 2539–2547, 2015.

Anders Boesen Lindbo Larsen, Søren Kaae Sønderby, Hugo Larochelle, and Ole Winther. Autoencoding beyond pixels using a learned similarity metric. *arXiv preprint arXiv:1512.09300*, 2015.

Tod S Levitt and Daryl T Lawton. Qualitative navigation for mobile robots. *Artificial intelligence*, 44(3):305–360, 1990.

Tong Lin and Hongbin Zha. Riemannian manifold learning. *IEEE Transactions on Pattern Analysis and Machine Intelligence*, 30(5):796–809, 2008.

Lars Mescheder, Sebastian Nowozin, and Andreas Geiger. Adversarial variational bayes: Unifying variational autoencoders and generative adversarial networks. *arXiv preprint arXiv:1701.04722*, 2017.

Nicholas Metropolis, Arianna W Rosenbluth, Marshall N Rosenbluth, Augusta H Teller, and Edward Teller. Equation of state calculations by fast computing machines. *The journal of chemical physics*, 21(6):1087–1092, 1953.

Jean-Arcady Meyer and David Filliat. Map-based navigation in mobile robots:: Ii. a review of map-learning and path-planning strategies. *Cognitive Systems Research*, 4(4):283–317, 2003.

Charlie Nash and Chris KI Williams. The shape variational autoencoder: A deep generative model of part-segmented 3d objects. In *Computer Graphics Forum*, volume 36, pages 1–12. Wiley Online Library, 2017.

Nasser M Nasrabadi. Pattern recognition and machine learning. *Journal of electronic imaging*, 16(4):049901, 2007.

Radford M Neal and Geoffrey E Hinton. A view of the em algorithm that justifies incremental, sparse, and other variants. In *Learning in graphical models*, pages 355–368. Springer, 1998.

John Paisley, David Blei, and Michael Jordan. Variational bayesian inference with stochastic search. *arXiv preprint arXiv:1206.6430*, 2012.

Giorgio Parisi. *Field theory, disorder and simulations*, volume 49. World Scientific, 1992.

Nicklas Persson. Shortest paths and geodesics in metric spaces, 2013.

Christopher Poultney, Sumit Chopra, Yann L Cun, et al. Efficient learning of sparse representations with an energy-based model. In *Advances in neural information processing systems*, pages 1137–1144, 2007.

Yunchen Pu, Zhe Gan, Ricardo Henao, Xin Yuan, Chunyuan Li, Andrew Stevens, and Lawrence Carin. Variational autoencoder for deep learning of images, labels and captions. In *Advances in neural information processing systems*, pages 2352–2360, 2016.

KATHERINE REDFIELD. Finding geodesics on surfaces. 2007.

Danilo Jimenez Rezende, Shakir Mohamed, and Daan Wierstra. Stochastic backpropagation and approximate inference in deep generative models. *arXiv preprint arXiv:1401.4082*, 2014.

Salah Rifai, Pascal Vincent, Xavier Muller, Xavier Glorot, and Yoshua Bengio. Contractive auto-encoders: Explicit invariance during feature extraction. In *Proceedings of the 28th International Conference on International Conference on Machine Learning*, pages 833–840. Omnipress, 2011.

Herbert Robbins and Sutton Monro. A stochastic approximation method. *The annals of mathematical statistics*, pages 400–407, 1951.

Mihaela Rosca, Balaji Lakshminarayanan, David Warde-Farley, and Shakir Mohamed. Variational approaches for auto-encoding generative adversarial networks. *arXiv preprint arXiv:1706.04987*, 2017.

Tim Salimans, David A Knowles, et al. Fixed-form variational posterior approximation through stochastic linear regression. *Bayesian Analysis*, 8(4):837–882, 2013.

José Santos-Victor, Giulio Sandini, Francesca Curotto, and Stefano Garibaldi. Divergent stereo for robot navigation: Learning from bees. In *Computer Vision and Pattern Recognition, 1993. Proceedings CVPR'93., 1993 IEEE Computer Society Conference on*, pages 434–439. IEEE, 1993.

Lawrence K Saul, Tommi Jaakkola, and Michael I Jordan. Mean field theory for sigmoid belief networks. *Journal of artificial intelligence research*, 4:61–76, 1996.

Stanislau Semeniuta, Aliaksei Severyn, and Erhardt Barth. A hybrid convolutional variational autoencoder for text generation. *arXiv preprint arXiv:1702.02390*, 2017.

Paul Smolensky. Information processing in dynamical systems: Foundations of harmony theory. Technical report, COLORADO UNIV AT BOULDER DEPT OF COMPUTER SCIENCE, 1986.

Pascal Vincent, Hugo Larochelle, Isabelle Lajoie, Yoshua Bengio, and Pierre-Antoine Manzagol. Stacked denoising autoencoders: Learning useful representations in a deep network with a local denoising criterion. *Journal of Machine Learning Research*, 11(Dec):3371–3408, 2010.

Martin J Wainwright, Michael I Jordan, et al. Graphical models, exponential families, and variational inference. *Foundations and Trends® in Machine Learning*, 1(1–2):1–305, 2008.

Wikipedia contributors. Discriminative model, 2004. URL `https://en.wikipedia.org/w/index.php?title=Discriminative_model&oldid=835201411`. [Online; accessed 20-April-2018].

Xinchen Yan, Jimei Yang, Kihyuk Sohn, and Honglak Lee. Attribute2image: Conditional image generation from visual attributes. In *European Conference on Computer Vision*, pages 776–791. Springer, 2016.